\documentclass[letterpaper, 10 pt, conference]{ieeeconf}  % Comment this line out if you need a4paper

\IEEEoverridecommandlockouts                              % This command is only needed if
\usepackage{graphicx} % for pdf, bitmapped graphics files
\usepackage{amsmath} % assumes amsmath package installed
\usepackage{amssymb}  % assumes amsmath package installed
\usepackage{pseudocode}
\usepackage[ruled,vlined]{algorithm2e}
\usepackage{color}    % use colors

\usepackage[caption=true, font=footnotesize]{subfig}

\usepackage{tabularx}
\usepackage{booktabs}
\usepackage{multirow}
\usepackage{adjustbox}

\newcommand{\wrt}{w.\,r.\,t.\ }

\definecolor{myblue}{rgb}{0.02,0.27,0.68}
\DeclareMathOperator*{\argmax}{arg\,max}

\title{
Planning with a Receding Horizon for Manipulation in Clutter \\
using a Learned Value Function
}

\author{Wissam Bejjani, Rafael Papallas, Matteo Leonetti and Mehmet R. Dogar% <-this % stops a space
		   \thanks{Authors are with the School of Computing, University of Leeds, United Kingdom
        {\tt\small  \{w.bejjani, r.papallas, m.leonetti, m.r.dogar\}@leeds.ac.uk}}%
}

\begin{document}

\maketitle
\thispagestyle{empty}
\pagestyle{empty}

%%%%%%%%%%%%%%%%%%%%%%%%%%%%%%%%%%%%%%%%%%%%%%%%%%%%%%%%%%%%%%%%%%%%%%%%%%%%%%%%
\begin{abstract}
Manipulation in clutter requires solving complex sequential decision making problems in an environment rich with physical interactions. The transfer of motion planning solutions from simulation to the real world, in open-loop, suffers from the inherent uncertainty in modelling real world physics. We propose interleaving planning and execution in real-time, in a closed-loop setting, using a Receding Horizon Planner (RHP) for pushing manipulation in clutter. In this context, we address the problem of finding a suitable value function based heuristic for efficient planning, and for estimating the cost-to-go from the horizon to the goal. We estimate such a value function first by using plans generated by an existing sampling-based planner. Then, we further optimize the value function through reinforcement learning. We evaluate our approach and compare it to state-of-the-art planning techniques for manipulation in clutter. We conduct experiments in simulation with artificially injected uncertainty on the physics parameters, as well as in real world tasks of manipulation in clutter. We show that this approach enables the robot to react to the uncertain dynamics of the real world effectively.

\end{abstract}

%%%%%%%%%%%%%%%%%%%%%%%%%%%%%%%%%%%%%%%%%%%%%%%%%%%%%%%%%%%%%%%%%%%%%%%%%%%%%%%%
\section{Introduction} \label{sec:intro}

We propose a planning approach for physics-based manipulation in clutter, for tasks in which an object has to be pushed to a goal region, with little to no repositioning of other objects.
Such robotic manipulation skills are required in a variety of applications.
In service robotics, for example, robots have to simultaneously interact with
multiple everyday objects (e.g., objects in drawers or in cabinets) to execute
household activities \cite{leidner2016robotic,dogar2012physics}. In industrial
settings, such as in warehouses, robots have to fulfill orders by picking
items off cluttered shelves, as in the Amazon Picking Challenge
\cite{hernandez2016team}. This requires pushing certain items out of the way,
without dropping them off the shelf, while reaching for a target item.

There has been significant recent interest in motion planning for pushing-based
manipulation tasks in clutter, and impressive planners have been proposed
\cite{haustein2015kinodynamic,kitaev2015physics,king2015nonprehensile}.  Real-world execution of these trajectories, however, still poses great challenges.
The main difficulty is due to the inevitable inaccuracy in the physics model used by
the planners. This inaccuracy is emphasized particularly when multiple objects
are in contact, which is common in the application domains mentioned above.

\begin{figure}[!t] %[thpb]
    \captionsetup[subfigure]{labelformat=empty}
    \subfloat[]{\adjustbox{margin=1em,width=0.08\textwidth,set height=0.1cm,angle=90}{Open-loop}}\hspace*{-0.9em}
    \subfloat[]{\adjustbox{margin=1em,width=0.12\textwidth,set height=0.13cm,angle=90}{Kino-dynamic planner}}\hspace*{-1.0em}
    \subfloat[]{ \includegraphics[width=0.235\columnwidth]{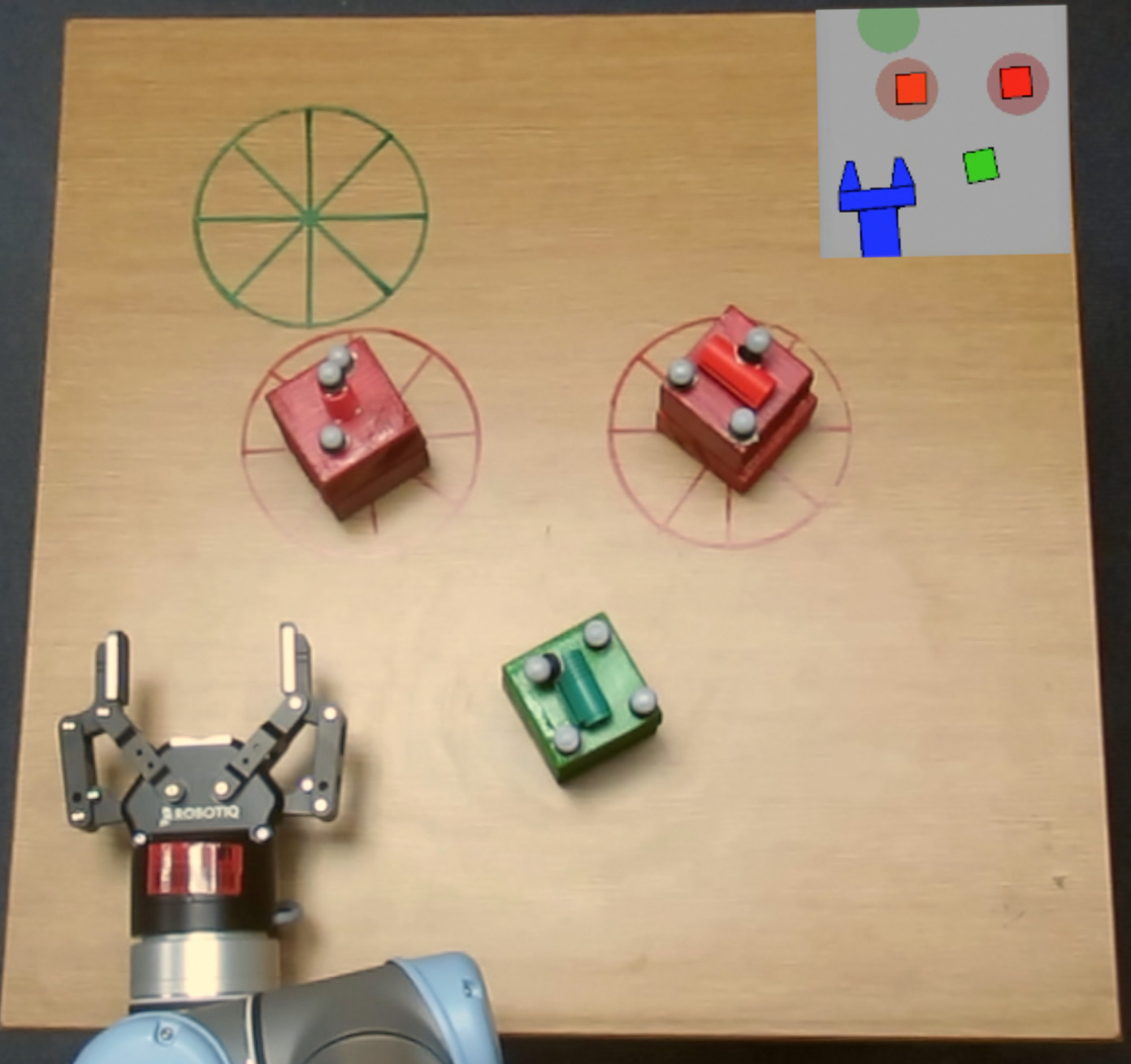}  }\hspace*{-0.92em}
    \subfloat[]{ \includegraphics[width=0.235\columnwidth]{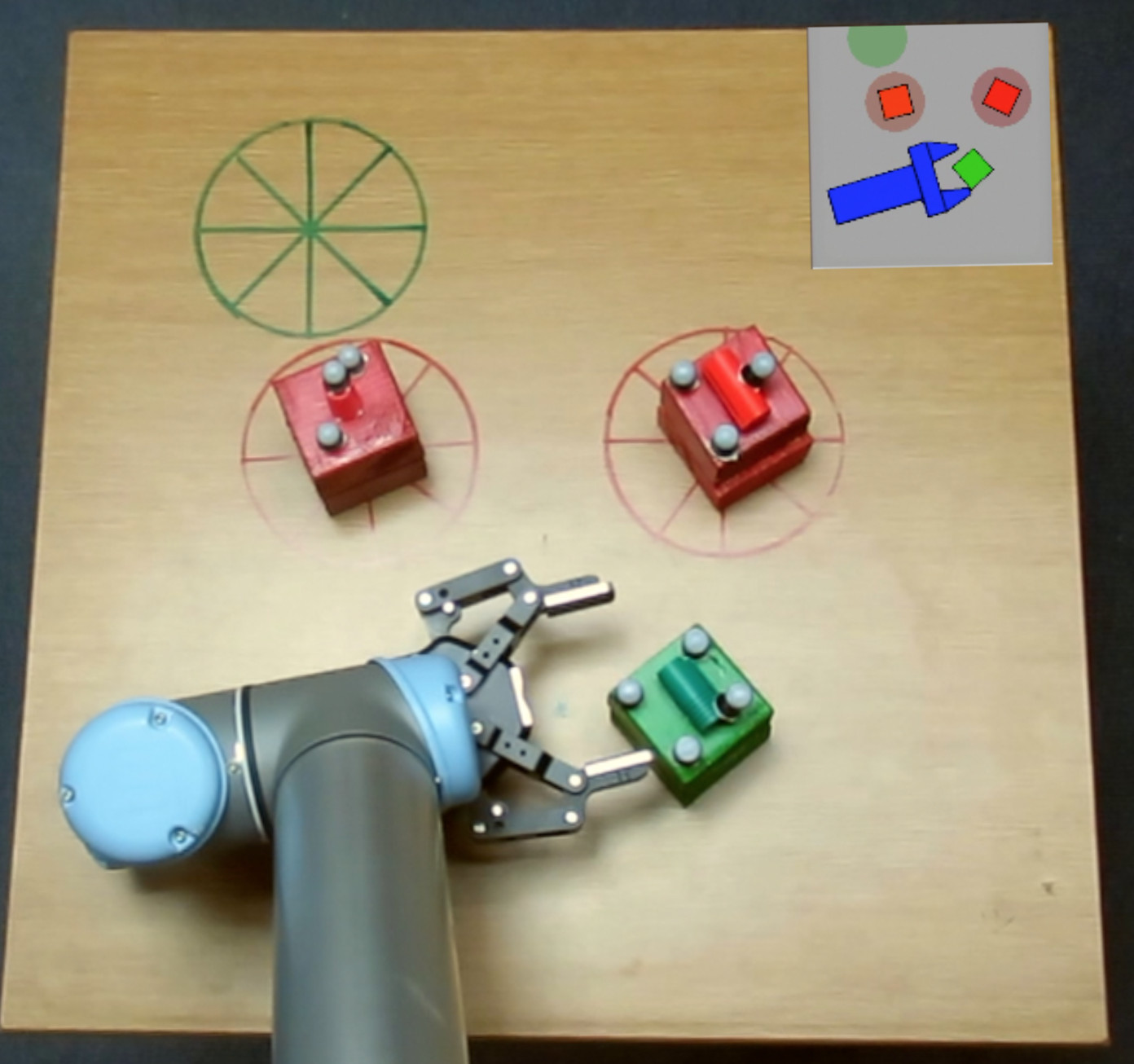} }\hspace*{-0.92em}
    \subfloat[]{ \includegraphics[width=0.235\columnwidth]{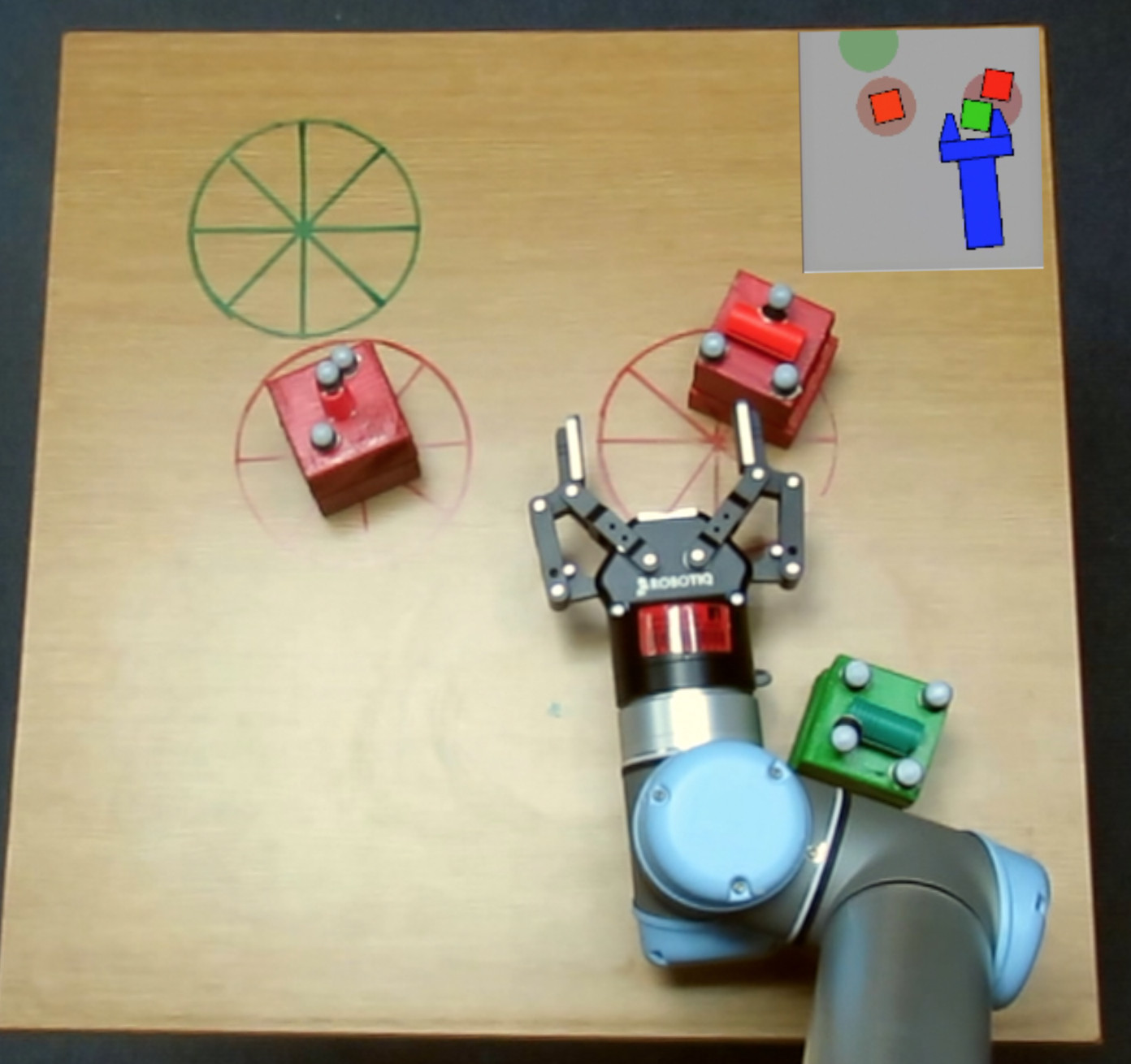}  }\hspace*{-0.92em}
    \subfloat[]{ \includegraphics[width=0.235\columnwidth]{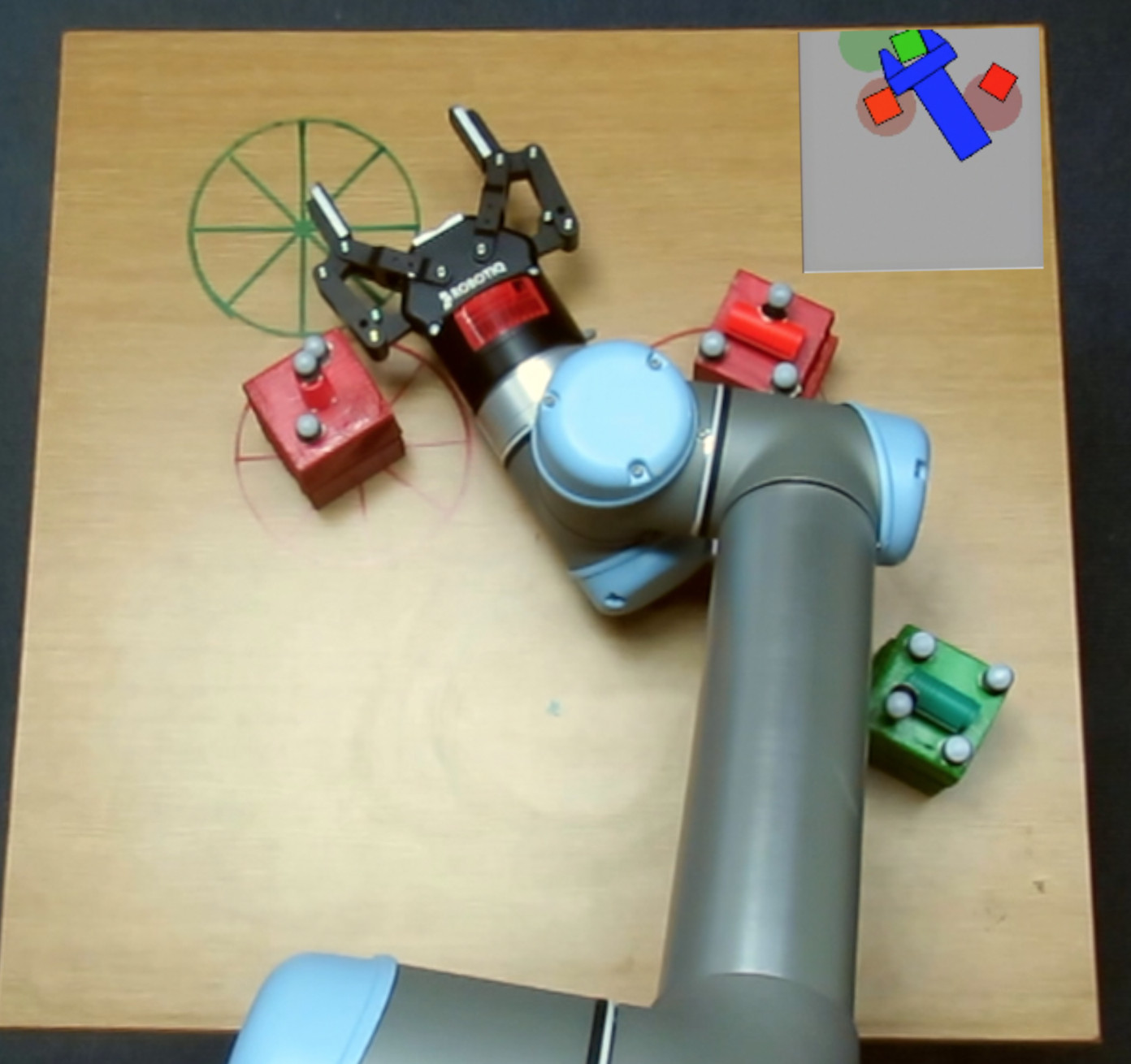}  }\hspace*{-0.92em} \\[-1ex]
    \subfloat[]{\adjustbox{margin=1em,width=0.08\textwidth,set height=0.1cm,angle=90}{Closed-loop}}\hspace*{-0.7em}
    \subfloat[]{\adjustbox{margin=1em,width=0.09\textwidth,set height=0.1cm,angle=90}{RHP execution}}\hspace*{-1.0em}
    \subfloat[]{ \includegraphics[width=0.235\columnwidth]{148po1c.jpeg}  }\hspace*{-0.92em}
    \subfloat[]{ \includegraphics[width=0.235\columnwidth]{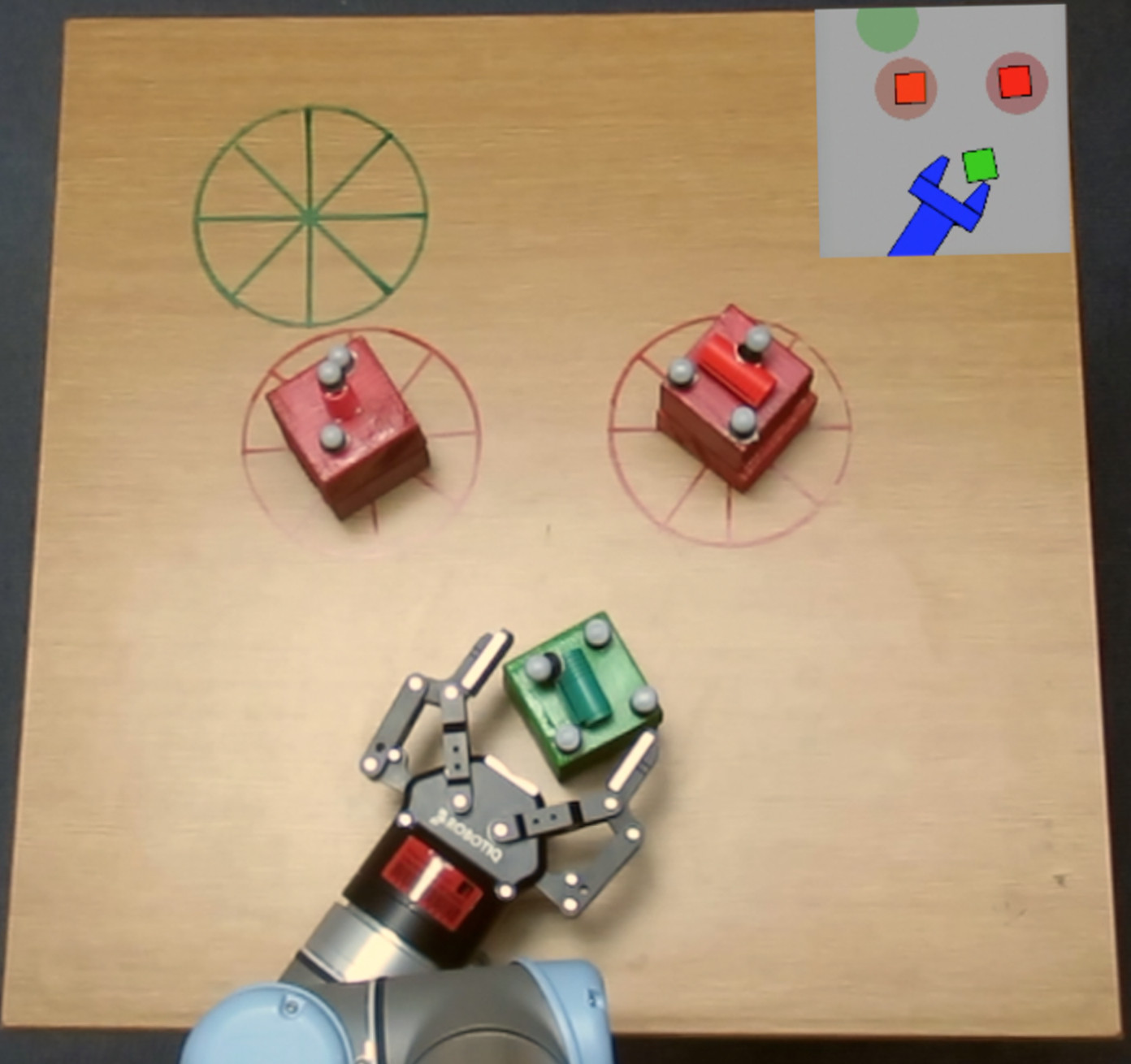}  }\hspace*{-0.92em}
    \subfloat[]{ \includegraphics[width=0.235\columnwidth]{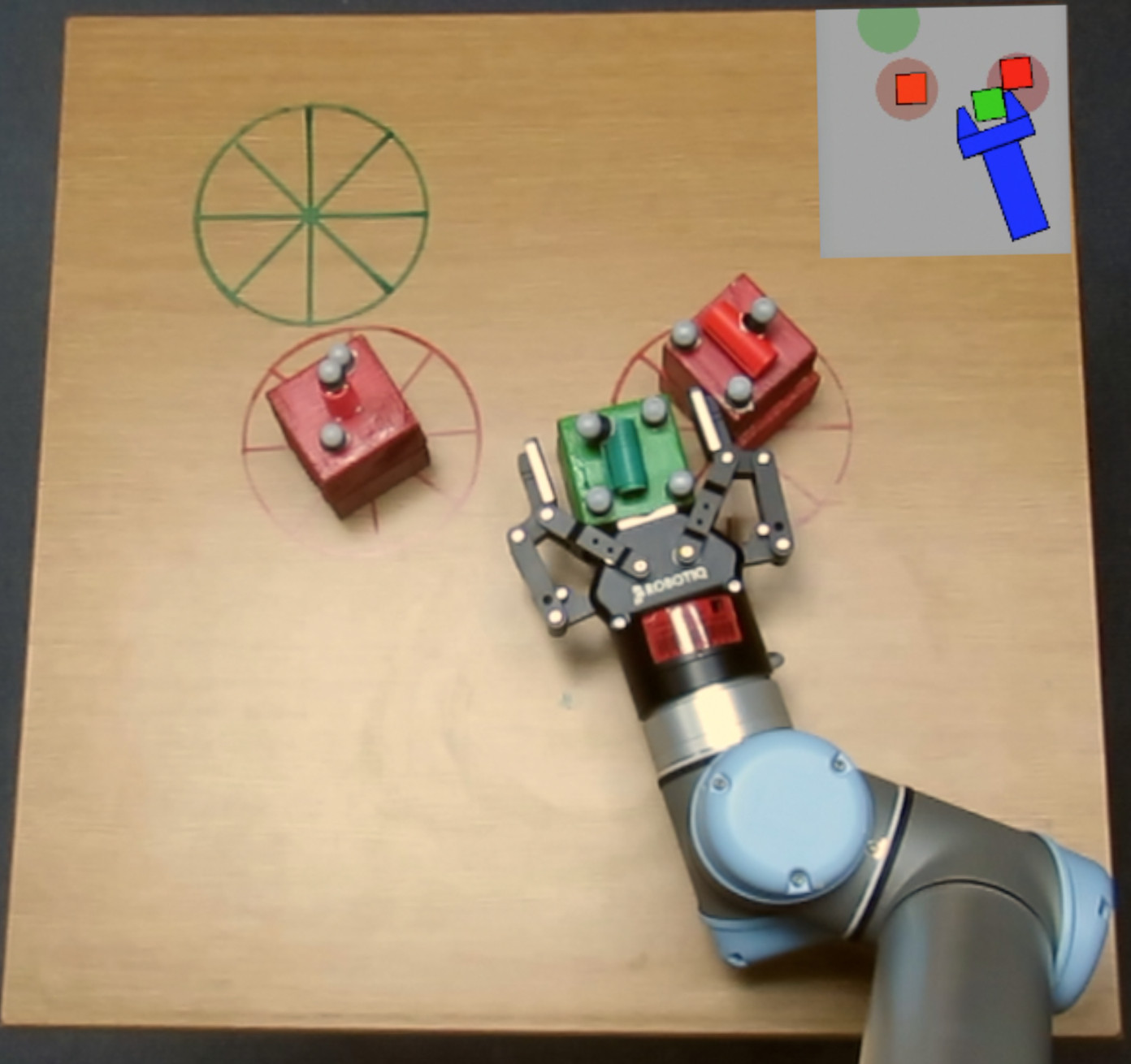}  }\hspace*{-0.92em}
    \subfloat[]{ \includegraphics[width=0.235\columnwidth]{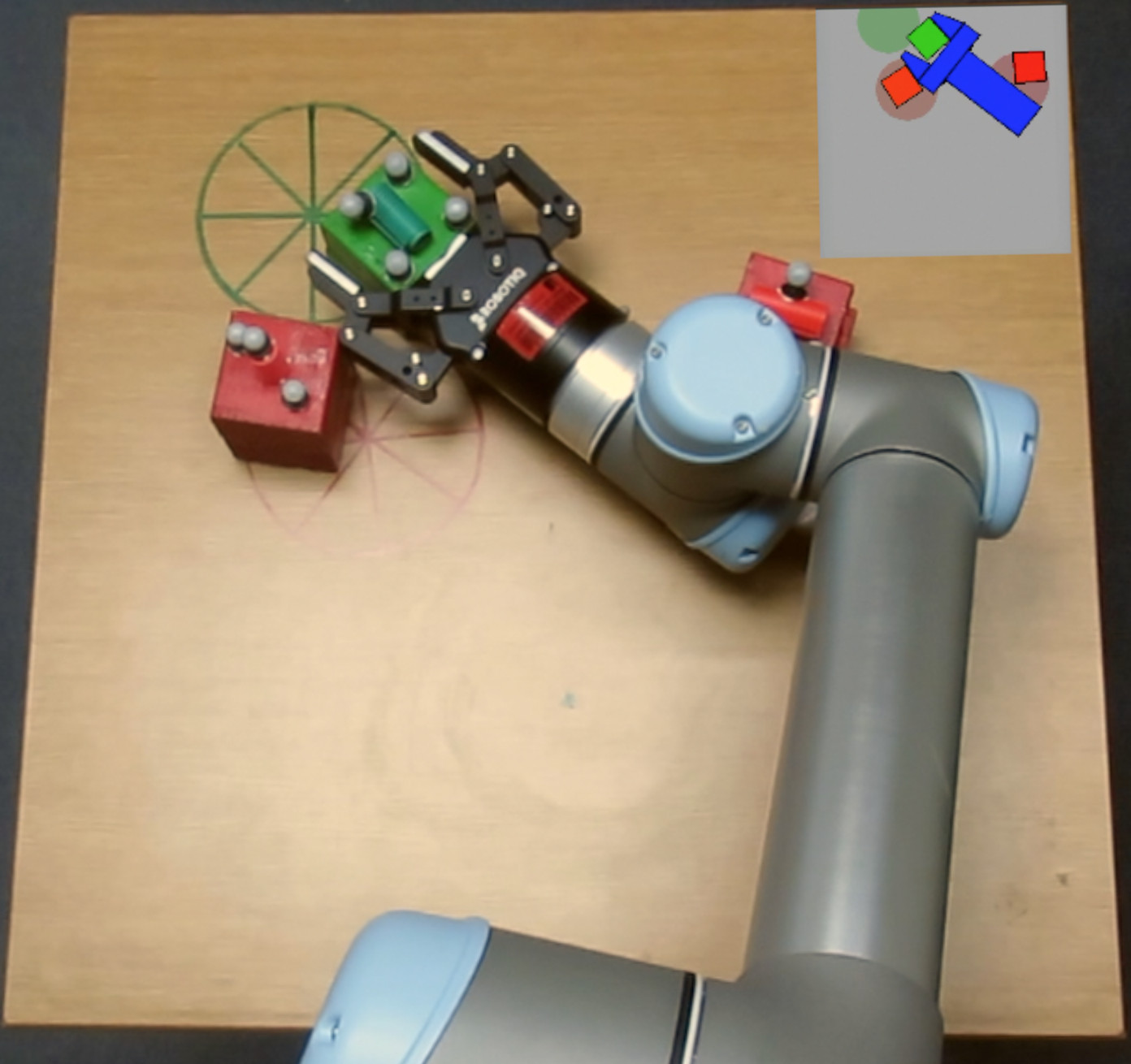}  }\hspace*{-0.92em}
    \caption{Top: robot failing to push the green box to the goal region by following a precomputed plan using kino-dynamic planning. Bottom: robot successfully executing the task using closed-loop RHP execution.}
    \label{fig:real_example_1}
\end{figure}

We present an example of the task in Fig.~\ref{fig:real_example_1}, where the green object has to be pushed to a target region (the green region) while keeping the red
objects close to their original positions (red regions).
The top row of Fig.~\ref{fig:real_example_1} shows the execution of an open-loop trajectory generated by a sampling-based planner, while the bottom row presents an execution of our system.
The
overlaid animated figures (on the top-right corner of the images) show the
planner's prediction of how the objects should move during interaction. When planned
trajectories are executed open-loop, the real motion
of the objects can differ significantly from the motion predicted by the
planner.  For this reason, in the shown example, the open-loop controller fails to
accomplish the task.

A solution to this problem is to interleave planning and execution. In this approach,
a sequence of actions is planned, but only the first action
in this sequence is executed. Then, the current state is updated by
observing the environment, after which another sequence of actions is
planned, and the routine is repeated. This idea is commonly used in domains
that involve uncertainty, and underlies many similar
methods with different names, among which: rolling horizon planning, receding horizon control, and model predictive
control.
We show an execution of such a controller in the bottom row of
Fig.~\ref{fig:real_example_1}. Even if objects move differently than predicted,
the controller has the opportunity to correct for it.

One possible approach to generating such a receding horizon planner (RHP) would be to run one of
the aforementioned planners at every step of the execution, to generate the new
sequence of actions.  The computation time these planners require, however, is
prohibitively high, typically taking from tens of seconds to minutes for
one plan
\cite{dogar2012planning,haustein2015kinodynamic,kitaev2015physics,king2015nonprehensile,dogar2012physics,johnson2016convergent,mahayuddin2017randomized}.
In contrast, we are interested in real-time execution, which requires a planner
that can quickly suggest an action for the current state of the world.

To generate plans quickly, we propose to run RHP with a short horizon into the
future, and take advantage of an appropriate cost-to-go function as a proxy for the \emph{value} of the states beyond the horizon. This
value of a state must estimate how costly (or rewarding, depending on the formulation) it would be to reach the goal from
that state. In domains where multiple physics-based object-to-object contact is possible,
defining this function is a challenge on its own.

\begin{figure}[!t]
    \centering
    \includegraphics[width=\columnwidth]{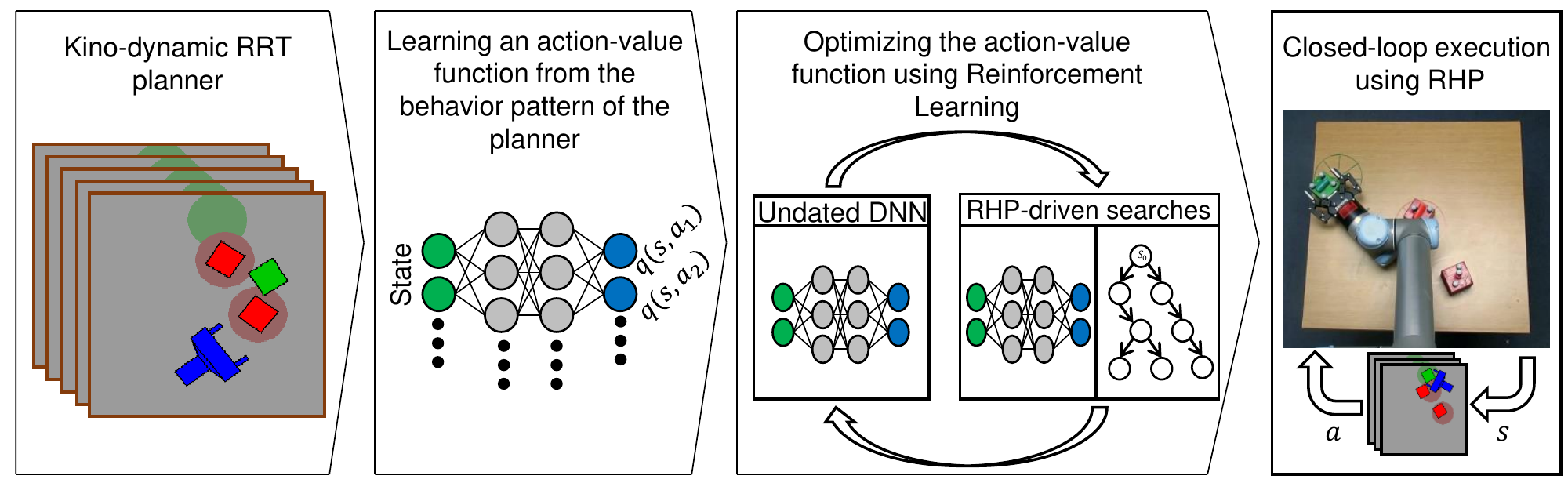}
    \caption{Overview of the proposed approach.}
    \label{fig:wf}
    \vspace{-2mm}
\end{figure}

In this paper, we propose to use data-driven techniques to learn such a
value function, so that we can then use it as a heuristic within a receding horizon
planner. In order to do this, in simulation, we generate many planning instances, and we
solve them using an existing sampling-based planner (a kino-dynamic RRT planner
\cite{haustein2015kinodynamic}). We then use the sequences of states and
actions in these plans to train a Deep Neural Network (DNN), to predict the
value of a state-action pair, that is, the expected reward for reaching the goal starting from that state and using that action.

The value function learned at this stage leads to an acceptable controller when used as a heuristic, but as we show in our
results, it can be further improved. Our insight is that the DNN,
trained only by the sampling-based planner, encodes the value of a state under the planned trajectory, which differs from what the robot will actually execute when controlled by RHP.
Therefore, we use a reinforcement learning (RL) algorithm to gradually update
the value function to better estimate the actual optimal value of the manipulation task.
% j

Our key contribution is in showing that such an approach gives promising
results in the domain of physics-based manipulation in clutter: the robot is able to
perform fast closed-loop re-planning to deal with the inherent uncertainty in this domain, which challenges existing planners.  We show
that a heuristic value function can be learned using sampling-based planners and
deep learning. Furthermore, we show that it can be improved using reinforcement learning.  We
perform simulated and real-world experiments to evaluate the performance of our
planner.

\section{Related Work} \label{sec:relwo}

The planners discussed so
far~\cite{haustein2015kinodynamic,kitaev2015physics,king2015nonprehensile,dogar2012physics}
adopt an approach of motion planning followed by open-loop execution to solve
the problem of manipulation in clutter.  In particular, Haustein et
al.~\cite{haustein2015kinodynamic} adopts sampling-based planning to solve this
problem. They propose reducing the search space of kinodynamic Rapidly
exploring Random Trees (RRT) by planning over statically stable environment
states while allowing for physical interaction in-between these states. In this
paper, we use a similar kinodynamic RRT planner to generate plans to different
manipulation problem instances.
There are planners which also take uncertainty into account before the
generation of the motion trajectory
\cite{dogar2012physics,koval2015robust,johnson2016convergent,mahayuddin2017randomized},
but these planners typically rely on uncertainty reducing actions which
generate a conservative sequence of actions, limiting the robot from using the
complete dynamics of the domain.

In this paper, however, we are mainly interested in real-time planning which can be used in a closed-loop system
to respond dynamically to changes during execution. Kloss et al.~\cite{kloss2017combining} present a learning approach for planar pushing tasks
in closed-loop form. They train a neural network, which takes the visual state
of the environment as input, and feeds the appropriate physical properties extracted from
the scene to an analytical model of the task. Hogan and Rodriguez
\cite{hogan2016feedback} apply a feedback control scheme that can alternate
between different interaction modes to control a tool pushing a slider
on a planar surface. To avoid learning a behavior that exploits the idiosyncrasies of the physics model in the simulation environment, Peng et al.~\cite{peng2017sim} propose randomizing the physics parameters in the simulation environment during the learning phase for a pushing task.
These approaches have proven capable in real world manipulation. However, they focus on manipulating a single object whereas we are interested in multi-object interaction present in cluttered spaces.
Laskey et al.~\cite{laskey2016robot} tackles this problem by relying on a expert human demonstrator and a DNN to control 2-DOF of a robot arm to reach a target object on a cluttered surface.
% In a similar task setting to ours,
% Laskey et al.~\cite{laskey2016robot} use a
% The DNN learns a policy from plans generated
% on-line using the Dataset Aggregation algorithm.
% They propose using a hierarchy of
% supervisors with different skill levels to alleviate the cost associated with
% on-line learning. This approach, however, requires access to a near optimal task supervisor.

The idea of learning a heuristic for control and planning has been
applied in domain other than manipulation in clutter.
Negenborn et al.~\cite{negenborn2005learning} propose a framework for learning based model predictive control for general Markov Decision Process. Similarly, Zhong et al.~\cite{zhong2013value} look at the problem of value function approximation for automatically shortening the horizon in model predictive control for dynamic tasks like inverted pendulum and acrobot. These approaches do not take into account challenges typical to clutter manipulation tasks, such as inaccuracy of the physics model, and computation constraints imposed by the physics engine.
Conceptually, the work of Anthony et al.~\cite{anthony2017thinking} is reminiscent to our work. They use Monte Carlo Tree Search to generate plans leading to the goal. They suggest making the searches more efficient by biasing the search process with a DNN-based value function that is recursively trained on previous iteration of the generated plans. Likewise, Hottung et al.~\cite{hottung2017deep} integrate neural networks in a heuristic tree search procedure for accelerating the searches by pruning the tree.
Further, Hussein et al. \cite{hussein2017deep} rely on collected demonstrations to pre-train a DNN-based policy. Then, they refine the policy in an active learning fashion where an agent is assumed to have access to the optimal policy when faced with states with low action confidence.
Although combining traditional control and theoretic planning such MCTS methods~\cite{browne2012survey,kocsis2006bandit} with machine learning offers promising solutions to problems with a sparse reward function \cite{silver2016mastering},
they are yet to be proven capable in handling physics based manipulation, where simulating a large number of roll-outs (a common feature of these approaches) at every time step is prohibitively expensive.
To the best of our knowledge, this line of thought has not yet been investigated in the context manipulation in clutter. In this paper we examine the problem from a physics based perspective for real world applications.
% In this paper we examine defining this loss function to the manipulation domain when learning a action-value function from sub-optimal sampling-based planners (Sec.~\ref{sec:IL}).

% Learning a policy that imitates demonstrations from an expert typically suffers from a limited dataset of transition samples, which only covers transitions on a path to the goal, while offering no information on alternative trajectories~ \cite{piot2014boosted}.
% In the recent work of Hester et al.~\cite{hester2017learning}, they jump-start a DNN-based policy with imitation learning before using it in reinforcement learning. To ensure that the policy learns from imitation a behavior that keeps the agent close to experienced part of the state-space, they impose a large margin classification loss on unexperienced state-actions pairs. Also, Lakshminarayanan et al.~\cite{lakshminarayanan2016reinforcement} achieve this using cross-entropy classification loss to favor transitions similar to one available in the demonstrations.

\section{Problem Formulation} \label{sec:RPPDE}
\begin{figure}[!t]
\centering
\includegraphics[scale=0.3]{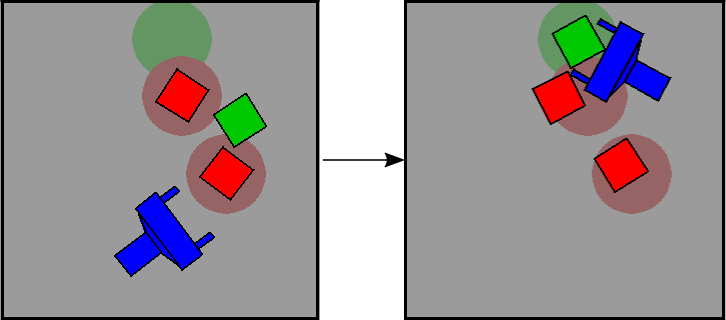}
\caption{The initial configuration (left) and the final configuration (right) of an example scene.}
\label{fig:manipTask}
\end{figure}

We target applications where a robot has to reach and manipulate objects in
cluttered spaces. For instance, reaching a ketchup bottle in the
back of a cluttered refrigerator shelf, or pushing a box item on a warehouse
shelf. In these settings, it is also desirable to minimize the disturbance of
other objects while manipulating the target object. The robot has to use
non-prehensile skills to manipulate movable objects from an initial state to a
desired goal state, as illustrated in Fig. \ref{fig:manipTask}.

We model the environment as a Markov Decision Process (MDP) represented as a
tuple ${ M = \langle S,A,T,r,\gamma \rangle }$ where $S$ and $A$ are the sets of
states and actions respectively,  ${ r : S \times S \rightarrow \mathbb{R} }$ is
the reward function, ${ T : S \times A \rightarrow S }$ is the transition function,
and $\gamma$ is the discount factor, such that ${ 0 \leq \gamma \leq 1 }$. Since we are interested in modeling planning problems with goal states, our MDP model is \emph{episodic}, that is there exists at least a (goal) state $s_g$ that can never be left ($T(s_g,a) = s_g \; \forall a \in A$) and gives zero reward ($r(s_g,s_g) = 0$).
A behavior for the MDP is represented as a (stochastic) \emph{policy} $\pi : S \times A \rightarrow [0, 1]$, where $\pi(s,a) = P(a | s)$, with $s \in S, a \in A$, that is, the probability of the agent picking action $a$ in state $s$. The \emph{value} of a state-action pair $\langle s, a \rangle$ under a given policy $\pi$ is the cumulative discounted reward (called the \emph{return}) achieved from $s$ taking $a$ and following $\pi$ thereafter: $q^\pi(s,a) = \sum_{t = 0}^{\infty} \gamma^t r_{t+1} |_{\pi}$\footnote{
% Note that this series converges, at least for some policies, under the assumption of the MDP being episodic.
Note that this series converges, at least for some policies,
% when $\gamma \leqslant 1$ for episodic MDP and $\gamma < 1$ for non-episodic MDP.
% for episodic MDP with $\gamma \leqslant 1$ and for non-episodic MDP with $\gamma < 1$.
under the assumption of $\gamma < 1$ for non-episodic MDP and $\gamma \leqslant 1$ for episodic MDP.
}

The state at time $t$ is given by the planar poses of the robot and the $m$ objects
$s_t = \{\langle x_{\textrm{robot}}, y_{\textrm{robot}}, \theta_{\textrm{robot}} \rangle,$ $ \langle x_{\textrm{obj\_}i}, y_{\textrm{obj\_}i}, \theta_{\textrm{obj\_}i} \rangle_{i=1}^m \}$.
%where the $R_{\{x,y,\theta\}}$ are the position and orientation of the
%end-effector in the world's frame of reference, $G_{i_{\{x,y\}}}^R$ and
%$O_{i_{\{x,y\}}}^R$, with $i \in \{1,\ldots, m\}$, are the centers of the goal
%region for object $i$, and the position of object $i$, with respect to the
%end-effector.
While the state space is continuous, we discretize the action space, so that
the robot can execute six actions in $A$, four to apply a force in each of the
cardinal directions, and two to rotate the end-effector clockwise and
counterclockwise in the task space.  The transition function $T$ models the
physics-based interaction between the robot and the objects, predicting how the
objects move in response to robot actions.
We avoid hand-crafted reward functions by setting
$r(s_t,s_{t+1}) = r_{t+1} = -1$ where $s_t,s_{t+1} \in S$. This
domain-independent reward function encourages the robot to reach the goal
${ S_{\textrm{goal}} \subseteq S }$ as fast as possible.

A \textit{task instance} $\langle s_{init}, G \rangle$ for our problem is defined by the initial positions of
the robot and objects, $s_{init}$, and a set of circular goal regions for all
objects ${ G = \langle x_{\textrm{goal\_}i}, y_{\textrm{goal\_}i},
radius_{\textrm{goal\_}i} \rangle_{i=1}^m }$, where $x_{\textrm{goal\_}i}$ and
$y_{\textrm{goal\_}i}$ is the centre of the goal region for object $i$ and the
$radius_{\textrm{goal\_}i}$ is the radius of this region. Example regions are
in Fig.~\ref{fig:manipTask}. In this paper, for the obstacle objects (red
objects in the figure), we place the goal regions (red regions in the figure)
on the initial positions of these objects at $s_0$. This is how we discourage
the planner from disturbing the scene. For the target
object (green object) the goal region (green region in the figure) can be anywhere.
We also use the same fixed radius for all goal regions.
The goal set ${ S_{\textrm{goal}} }$ is the set of all the states where all
the objects are in their goal regions $G$.

A \textit{plan} is a sequence of states and actions
${ p = \langle s_0, a_0, \ldots, s_{L-1}, a_{L-1} \rangle } $, where $L$ is the
length of the plan, $s_{t+1} = T(s_{t},a_t)$ with $s_t,s_{t+1} \in S$, and
$T(s_{L-1}, a_{L-1}) \in S_g$, i.e. the final state is a goal state.
An optimal plan is one that maximizes the return from the initial state.

Existing planners
\cite{dogar2012physics,haustein2015kinodynamic,kitaev2015physics,king2015nonprehensile}
can generate solutions to this problem (which are usually sub-optimal in the sense
of maximizing the return, but can at least find a plan to reach a goal state).
However, open-loop execution of these plans in the real world can easily fail,
as the predicted motion of the objects differs from the real motion due to
uncertainty in physics-based predictions, especially when there are multiple
contacts between objects.

In order to take the model uncertainty into account, we propose to interleave planning and execution, where the robot:
\begin{enumerate}
    \item Plans a sequence of actions from the current state $s_0$.
    \item Executes the first action, $a_0$, in the sequence.
    \item Observes the state of the system and update $s_0$.
    \item Goes to step 1.
\end{enumerate}
The first step in this procedure is usually the most costly one. The
planners in the literature addressing the problem of manipulation in clutter are
not fast enough (taking anywhere from tens of seconds to minutes
\cite{dogar2012physics,haustein2015kinodynamic,kitaev2015physics,king2015nonprehensile})
to run as step 1. Therefore, in this paper, we propose to plan for only a short
horizon, and not necessarily until the goal. Given
such a horizon $h$, we are then interested in maximizing the reward for $h$
steps plus our estimate of how much reward we can get from the state at the horizon:
% \begin{equation} \label{eq:n_step_cost}
%     \langle a_0, \ldots, a_{h-1} \rangle =
%     \argmax_{\langle a_0, \ldots, a_{h-1} \rangle} \sum_{k = 0}^{h-1}
%     \gamma^k r_{k+1} + \gamma^h  v(s_{h})
% \end{equation}
\begin{equation} \label{eq:n_step_cost}
    \langle a_0, \ldots, a_{h-1} \rangle =
    \argmax_{\langle a_0, \ldots, a_{h-1} \rangle} \sum_{k = 0}^{h-1}
    \gamma^k r_{k+1} + \gamma^h  \max_a q(s_{h},a)
\end{equation}
where the action-value function $q(s_{h},a)$ estimates the expected return for reaching the goal from the horizon state $s_h$ and choosing the action $a$.
In this paper, we call this the \textit{Receding horizon planner} (RHP) and we use it as the robot control policy.

The horizon can mitigate the inaccuracy of the value function estimate, by ranging from infinity, with the robot planning all the way to the goal, to zero, with the robot acting greedily with the respect to the value function. With an infinite horizon the value function is ignored, while with $h=0$ the behaviour depends entirely on the value function. In the latter case, if the value function is optimal (that is $q(s,a) \geq q^\pi(s,a) \; \forall s \in S, a \in A$, for any policy $\pi$), so is the resulting policy. In practice, a short but non-zero horizon takes advantage of both the planner and the value function without relying on either one entirely, and we experiment with several values for $h$.

% then maximizing the return above makes RHP optimal.
% The better is the estimate of $q$ to $q^*$ the shorter RHP horizon can be made.
% In this paper,
% we use learning techniques to learn a good estimate $\hat{q}$ of the optimal action-value function $q^*$ where  $q^* \geq \hat{q}$.
% The action-value is an estimate of the expected return starting from state $s$ taking action $a$.
% Once the action-value function is learned, one can substitute in Equation \ref{eq:n_step_cost} $v$ by $\max_aq$.

% A behavior for an MDP is represented as a (stochastic) \emph{policy} $\pi : S \times A \rightarrow [0, 1]$, where $\pi(s,a) = P(a | s)$, that is, the probability of the agent picking action $a$ in state $s$. If $\pi(s,a) = 1$ for some action $a$ in each state $s$ then the policy is deterministic and we denote it as $\pi(s) = a$. The action-value function of a given policy $\pi$ is defined as $q^\pi(s,a) = \mathop{\mathbb{E}}[ \sum_{t=0} \gamma^t r_{t+1}]$, which is the expected value of the sum of the discounted future rewards (called the \emph{return}). One way to compute an optimal policy common to planning and learning methods is to estimate the optimal action-value function, that is the function such that:  $q^*(s,a) \geq q^\pi(s,a) \; \forall s \in S, a \in A$. The optimal policy can then be retrieved by choosing, in each state, the action with the highest expected return: $\pi^*(s) =  \max_{a} q^*(s,a)$.

\section{Learning an action-value function from sampling-based planners}
We are interested in having an approximation of the action-value function over the entire continuous state-space for any goal.
In this section, we present a learning approach to extract an estimate of the action-value function from a collection of planning instances. The learned function approximates the return achieved by the planner from a number of sampled initial states and goal regions.

\subsection{Generating example plans} \label{sec:RRT}
Sampling-based planners treat every new planning instance independently from previously solved instances.
Also, they must plan until the goal, which means that they do not offer useful information on the searched areas of the state-state space from which the goal was not reached.
However, sampling-based planners provide a probabilistically complete tool to solve complex planning problems in high-dimensional state spaces without necessarily requiring a hand-crafted or domain-dependent heuristic.
In particular, Kino-dynamic planners are one family of the sampling-based Rapidly exploring Random Trees  planners, specific for solving planning problems that involve dynamic interactions.
% which are used when the control space is different from the state space.
We implement a state-of-the-art kino-dynamic planner~\cite{haustein2015kinodynamic} used for solving
physics-based manipulation in clutter planning problems.
%This kino-dynamic planner expand the RRT by sampling static state and allows for dynamic transitions between states.
We generate $P$ random problem instances
$\langle s_{init}^p, G^p \rangle_{p=1}^P$, as described in Sec.~\ref{sec:RPPDE}.
Then, for each planning instance $p$, we run the kino-dynamic planner to generate a solution of the form $\langle a_{init}^p, \ldots, a_{L-1}^p \rangle$.
The state trajectory $\langle s_{init}^p, \ldots, s_{L-1}^p \rangle$ induced by the action sequence of a plan $p$ brings the environment to a goal state $T(s_{L-1}^p, a_{L-1}^p) \in S_g^p$ where each box is placed in its corresponding target region $G^p$.

\subsection{Learning The Action-Value Function From Observed Trajectories}
\label{sec:IL}
We use example plans to train a DNN to predict the action-value estimate for a given state-action pair.
We represent the action-value function estimate $\hat{q}(s,a; \theta)$, modeled by a DNN with parameters $\theta$.
To train the DNN, we use every state-action pair encountered along every example plan.
For each example plan, and for every state-action pair in that plan, we compute the update target:
\begin{equation} \label{eq:q_tar_selected}
q(s_l^p,a_l^p) = \sum_{k = 0}^{L-l-1}  \gamma^k r_{l+k+1} =  r(\frac{1-\gamma^{L-l}}{1-\gamma})
\end{equation}
where $p$ stands for the index of the plan generated by the kino-dynamic planner and $l$ is the index of the state-action pair in that plan.
The second equality takes advantage of the fact that in our formulation all the immediate rewards, denoted as $r$, are the same\footnote{If $\gamma = 1$ the equation collapses to $q(s_l^p,a_l^p) = (L-l)r$}.

%\subsection{Grounding Unseen State-Action Pairs} \label{sec:GUSAP}

While the DNN trained as above learns to predict the action-values for the actions executed in a state by the kino-dynamic planner, the values predicted by the DNN for actions that have \textbf{not} been used by the planner can be arbitrary.
This is because the available example plans offer no information on the actions
that the planner did not choose along the traversed states.
As a result of function approximation, however,
these actions will nonetheless have a value.
The value can converge to an arbitrary number,
determined by the effect of the target value
in the states that the planner did traverse.
A possible undesirable effect is that
the values of the actions not chosen by the planner
can be higher than the chosen one.
This can later cause an action that was not
favored by the kino-dynamic planner to look more
favorable to RHP that uses the action-value
function as a heuristic.

In order to avoid this phenomenon, we ground
the unchosen actions to a target value that is
lower than the target value of the chosen action.
In literature~\cite{hester2017learning,lakshminarayanan2016reinforcement},
the difference between the value
of a desired action and the other actions is referred to as
the \emph{value margin}
% We will refer to the difference between the value
% of the chosen action and the other actions as
% the \emph{value margin}.

We propose a definition of the value margin driven
by the observation that, in the domain of pushing tasks,
a mistake is in most cases not irreparable,
but can be overcome through a number
of $k$ additional actions.
Hence, we use for the action-value of the unchosen actions the following update target:
\begin{equation}\label{eq:q_tar_unselected}
\begin{aligned}
q(s_l^p,a_{u}^p) =
\begin{cases}
r(\frac{1-\gamma^{L-l+k}}{1-\gamma}), \, & \text{if} \ \hat{q}(s_l^p,a_u^p;\theta) \geq q(s_l^p,a_l^p) \\
\hat{q}(s_l^p,a_{u}^p;\theta), \, & \text{otherwise}
\end{cases}
\end{aligned}
\end{equation}

%\quad \forall a \neq a_s \\
where $a_{u} \in A\setminus \{a_l\}$ is an unchosen action\footnote{if $\gamma = 1$ the first component of the equation collapses to $q(s_l^p,a_{u}^p) = (L-l+k)r$ }. This imposes that the unchosen actions, which would otherwise have a higher value than the chosen one, have a value equivalent to being $k$ steps further away from the goal than the chosen action.
If, on the other hand, the value that the approximator converges to does not favor an unchosen action then we leave it unchanged (as estimated by the network).
Lastly, we add an $L_2$ regularization term to the target function of Eq.~\ref{eq:q_tar_selected} and \ref{eq:q_tar_unselected}, to avoid over-fitting on the available plans.
%We
Once the training converges, we can use the action-value function to derive RHP policy for the robot, as described in Sec.~\ref{sec:RPPDE}.
%In

We experimented with DNNs of different sizes
and expressive power, but none could reliably
represent the behavior of the planner over a
large number of task instances.
We demonstrate experimentally that executing a greedy policy directly
on the output of the network
leads to a success rate much lower than using the
action-value function as heuristic for RHP policy, i.e. using the action-value function only after a few lookahead steps. Nonetheless, we show in the next section that the information compiled in the action-value function can be further optimized to play
a valuable role when used as a heuristic to drive RHP.

\section{Heuristic-Guided Deep Reinforcement Learning} \label{sec:SLAPtoSLAPRL}
The performance of action-value based RHP
is bounded by the quality of its heuristic.
So far, the knowledge encapsulated in the
action-value function has two shortcomings:
first, the plans generated by the kino-dynamic planner are,
in general, sub-optimal; and second, information is lost in the
approximation by the DNN, with consequent performance degradation
with respect to the kino-dynamic planner.
Furthermore, the action-value function estimates
the return based on the average behavior of
the kino-dynamic planner.
However, the RHP policy, that is actually controlling
the robot, can differ from the behavior of the
kino-dynamic planner.
To overcome these problems, we use reinforcement learning
to 1) improve the action-value function to better estimate
the optimal one and to 2) ground the unexperienced
state-space transitions to their actual values.
%S

We implement the Deep Q-Learning (DQN) algorithm \cite{mnih2013playing}.
We initialize the DNN to the trained DNN from the previous section.
% \footnote{Although feasible, learning a suitable action-value function for RHP using RL from scratch is extremely data inefficient and highly sensible to the hyper parameters of the training algorithm. By having a warm started DNN one can efficiently experiment with different RL algorithms with different hyper parameters.}.
Further, We formulate an RL policy, that we call
$\epsilon$-RHP, which selects a random action with
probability $\epsilon$ and with probability $1-\epsilon$
the policy queries RHP for an action.
We found that focusing the search towards the goal by
augmenting the RL policy with RHP,
reduces the chances of the action-value function from
diverging which is common problem in RL when used in
conjunction with a DNN as a function approximator.

Throughout the RL training process, the robot stores the
newly collected transition samples in a finite buffer
$D^{replay}$, initialized with transition samples from the
previously solved task instances, and gradually
replacing old samples.
At every action step, the DNN parameters are updated by
minimizing a loss function on a batch of random transition
samples from $D^{replay}$. The loss function is the squared
prediction error over the $M$ samples in a batch $B = \{ \langle s_i, a_i, r_{i}, s'_{i} \rangle_i\}$ where $s'_i$ is the state following $s_i$ in sample $i$:

\begin{equation}
L_{\theta}(B) = \sum_{i=1}^{M} (r_{i} + \gamma \max_{a'_{i}} \hat{q}(s'_{i},a'_{i} ; \theta) - \hat{q}(s_{i},a_{i}; \theta))^2
\end{equation}
We also add an $L_2$ regularization loss on the network parameters.

The benefit of $D^{replay}$ is twofold. It is using the collected experience more effectively to counteract the high correlation in the on-line incoming samples, which is also known as \emph{experience replay}. Second, it leads to a smooth change in the action-value function, and consequently in the robot behavior, as it shifts from estimating the cost of following the kino-dynamic planner to estimating the optimal action-value function.

\section{Searching the action-space up to the horizon}\label{sec:SLAP}
%\commentw{should we replace $s_0$ by $s_i$ or $s_t$ in this section to make it clearer?}
%The highly non-linear nature of the dynamical
%interactions between physical objects in a
%cluttered space makes it particularly hard for a
%function approximator to model the optimal
%action-value function $q^*$ using a reasonable amount
%of data and computation resources.
%Nevertheless, the action-value function $\hat{q}$ learned
%over the data from the kino-dynamic planner and
%RL stands as descent estimate of the optimal one.

%The low fidelity of physics models in cluttered environments and the near real-time execution %constraints dictate planning to a short horizon (otherwise the real world would diverge from the %expected state trajectory in simulation) and a small number of trajectory searches respectively.

We use the learned action-value function $\hat{q}(s,a;\theta)$ as
heuristic to RHP, i.e. we use it as an approximation in
Eq.~\ref{eq:n_step_cost} in-place of the unknown optimal action-value function at $s_h$.

The only remaining problem is searching the action-space up to the horizon, i.e. the maximization over the $h$ actions in Eq.~\ref{eq:n_step_cost}.
One na\"ive way to do this is to explore all possible action sequences up to the horizon $h$.
%Then execute the first action of the trajectory which maximizes the expected discounted sum of future rewards (see Equation \ref{eq:n_step_cost}).
However, an exhaustive search would scale badly with the horizon depth $h$ and the size of the action set $A$, $\mathcal{O}(|A|^h)$.

Instead, we bias the search towards promising actions. We implement RHP, such that it simulates $n$ trajectories, which we call roll-outs, of horizon $h$ each. Each of the $n$ roll-outs is started from the current state $s_0$. At every step $t$ in a roll-out, RHP samples an action using the soft-max of the action-value function:
\begin{equation}\label{SM}
P(a|s_t) = \frac{  exp(\hat{q}(s_t,a; \theta)/\tau)  }    {\Sigma_{a_i \in A} exp(\hat{q}(s_t,a_i; \theta)/\tau) }
\end{equation}
where $\tau$ is the temperature parameter. This would favor exploring actions whose value learned in the previous section is the highest.
%RHP then uses the model (simulation environment) to execute the action and access the next state and the immediate reward.
% When reaching the horizon $h$ or a goal state, the environment is reset to the roll-out start state $s_{0}$
% and
The return of a roll-out is computed as an $h$-step return,
where the first $h$ rewards are generated by the model,
and the action-value function acts as a proxy for the rewards beyond the horizon:
$$
R_{0:h} =  r_1 +  \gamma r_2 + \ldots + \gamma^{h-1}r_{h} + \gamma^{h}\hat{q}(s_{h}, a_h; \theta).
$$
%The $h$-step return is estimated by substituting the unknown optimal action-value $q^*$ with its approximation $\hat{q}$.
RHP then executes the first action in the roll-out that obtains the highest return.
This procedure reduces the number of simulated actions per RHP query to $n \times h$.
The action-value function, therefore, plays two roles:
to inform the search through the soft-max sampling, and as a proxy for the rewards that are not sampled from the model.

% At this stage the only difference in value between the actions that can be exploited by soft-max is the result of the target action-value in Eq. \ref{eq:q_tar_unselected}. This only guarantees that the policy induced by the planner emerges from the action-value function, but carries little information on the real value of the actions that the planner does not choose. The last phase of our method refines the learned action-value function through reinforcement learning, with the result of: optimizing the policy, making the behavior more robust, and providing a more accurate estimate of the unselected actions.

\begin{figure}[t]
\centering
% \framebox{\parbox{3in}{a2.png}}
\includegraphics[scale=0.5]{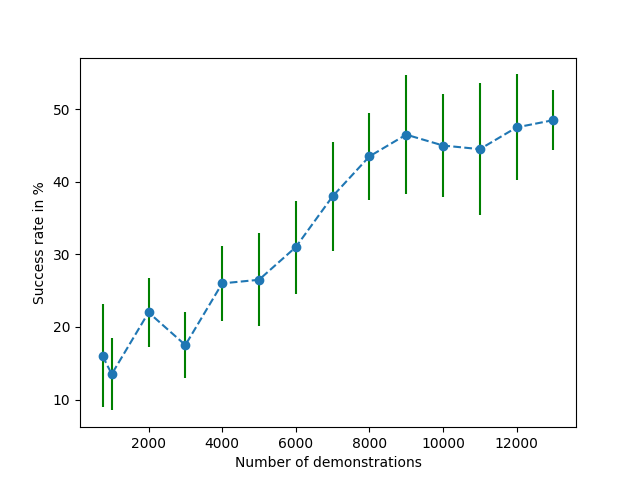}
\caption{Performance of the greedy policy induced by the action-value function trained over a different number plans.}
\label{fig:GPIL}
\end{figure}

\section{Evaluation} \label{sec:eval}
%W

The manipulation scenario follows the same model introduced in Sec.~\ref{sec:RPPDE}. As shown in Fig. \ref{fig:manipTask}, the environment consists of the end-effector of the robot arm shown in blue and 3 movable square boxes of side $6 \ cm$ on a planar surface shown in red and green. The robot has to push one of the boxes into a desired target region of radius $6 \ cm$, while having the rest of the objects placed by the end of the task as close as possible to their initial pose. The target regions are depicted by light colored circles corresponding to their designated boxes.
% A set of 500 random task instances is generated for evaluation purposes.

We evaluate the proposed approach in three experiment sets:
\begin{itemize}
    \item First, we measure \textit{the effect of the number of plans}
    (that is, plans generated by the kino-dynamic planner) on the
    quality of the action-value function.
    We show that after a certain number of plans from the planner, the
    performance of the induced policy hits a plateau.
    \item Second, we compare the performance of our trained planner to different base line approaches.
%     we analyze the contribution of the different components of the system, namely grounded imitation learning (GIL),  SLAP, and guided RL. We show that each of these components contribute significantly to achieving a high performance in the task. We also compare the system's performance against a state-of-the-art kinodynamic planner.
    \item Lastly, we demonstrate some example plans on a real world implementation.
%     the learned policy working on a real robot system.
\end{itemize}

%T

In our simulation experiments, we modeled the world in the Box2D physics simulator \cite{box2D}.
The robot motion is generated by applying momentary forces to its end-effector,
and waiting until the robot and objects came to a stop due to frictional
damping forces.
With this force, we observed that each translational action moved
the hand for a distance of around $5 \ cm$, and each rotational action moved the
hand for about $30^{o}$.
%W
The kino-dynamic planner typically needed 20-30 actions to reach the goal in
our experiments in a wide variety of task instances. Therefore, we set 40 actions as
the limit before we stopped our policy.
At the end of a run, if any the boxes was out of its corresponding target region, then we considered that run a failure.  Otherwise, we considered it a success.

\begin{figure}[t] %[thpb]
    \captionsetup[subfigure]{labelformat=empty}
    \setlength{\belowcaptionskip}{-10pt}
    \centering
    \subfloat[Kinodynamic Planner]{ \includegraphics[width=0.45\columnwidth]{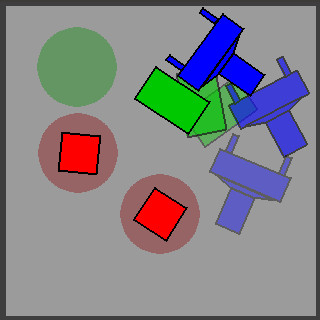}  }%\hspace*{-0.78em}
    \subfloat[RHP]{ \includegraphics[width=0.45\columnwidth]{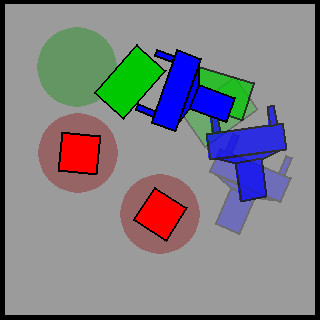} }\hspace*{-0.4em}
%     \hfill
    \caption{Evaluation at execution time with shape uncertainty.}
    \label{fig:sim_exp}
\end{figure}

\subsection{The effect of the dataset size on performance} \label{sec:exp1}
We generated multiple task instances by randomly
sampling non-colliding initial object poses and also randomly sampling a target
region for the green box.  We then run the kino-dynamic planner to generate a plan for each the task instances.
We used the TensorFlow
\cite{tensorflow2015-whitepaper} library to build and train a feed-forward
DNN model consisting of 5 fully connected layers.  The first 4
layers have 330, 180, 80, and 64 neurons, respectively, with ReLU activation
function. The output layer consists of 6 neurons, one per action, with
linear activation functions. The DNN is trained following the procedure described in Sec.~\ref{sec:IL} with the \emph{value margin} parametrized by $k=4$.

To test the action-value function encoded by the network, we generated 500 random task instances, and run a greedy  policy on them, that is executing the action with the highest value estimate.
% with horizon depth $h=1$ and roll-out $n=1$.
The horizontal axis in Fig.~\ref{fig:GPIL} shows the number of plans $P$ generated by the kino-dynamic planner, and
the vertical axis shows the success rate of the policy trained with that many plans.
We split the 500 task instances into 25 batches of 20, and the figure plots the average success rate and the confidence
interval over these batches.

As expected, the graph shows an increasing trend \wrt the number of
available plans. After reaching a $P$ of 9000 plans, we
see that it starts to plateau before it hits $50\%$ success rate. This demonstrates that the DNN alone could not encode, across all instances, a behavior as good as the planner which achieves a success rate of $98\%$ as shown in the first cell of Table~\ref{tab:results}.

\begin{figure}[t] %[thpb]
    \captionsetup[subfigure]{labelformat=empty}
    \setlength{\belowcaptionskip}{-10pt}
%     \centering
    \subfloat[]{\adjustbox{margin=1em,width=0.08\textwidth,set height=0.1cm,angle=90}{Open-loop}}\hspace*{-0.9em}
    \subfloat[]{\adjustbox{margin=1em,width=0.12\textwidth,set height=0.13cm,angle=90}{Kino-dynamic planner}}\hspace*{-1.0em}
    \subfloat[]{ \includegraphics[width=0.235\columnwidth]{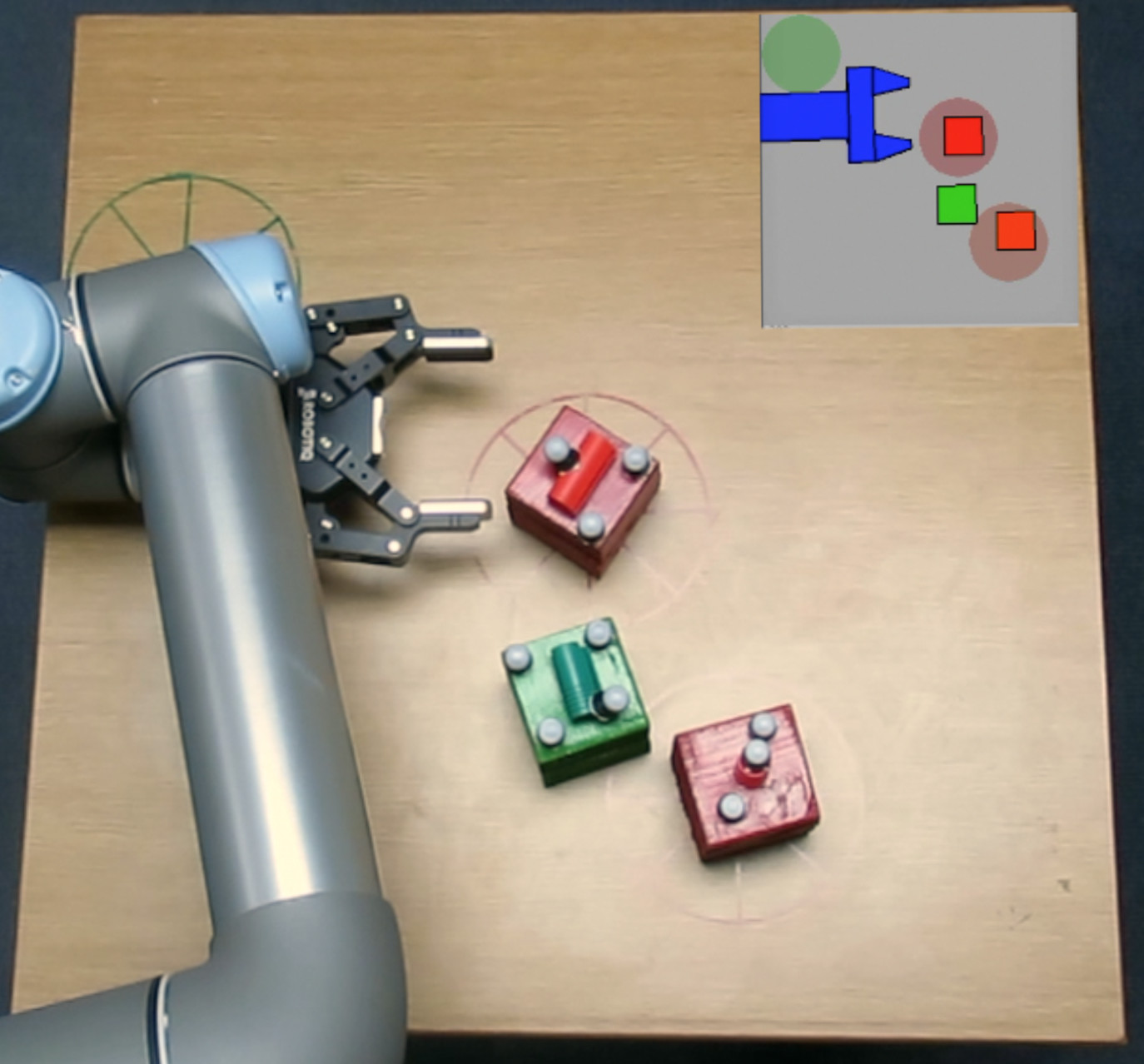}  }\hspace*{-0.92em}
    \subfloat[]{ \includegraphics[width=0.235\columnwidth]{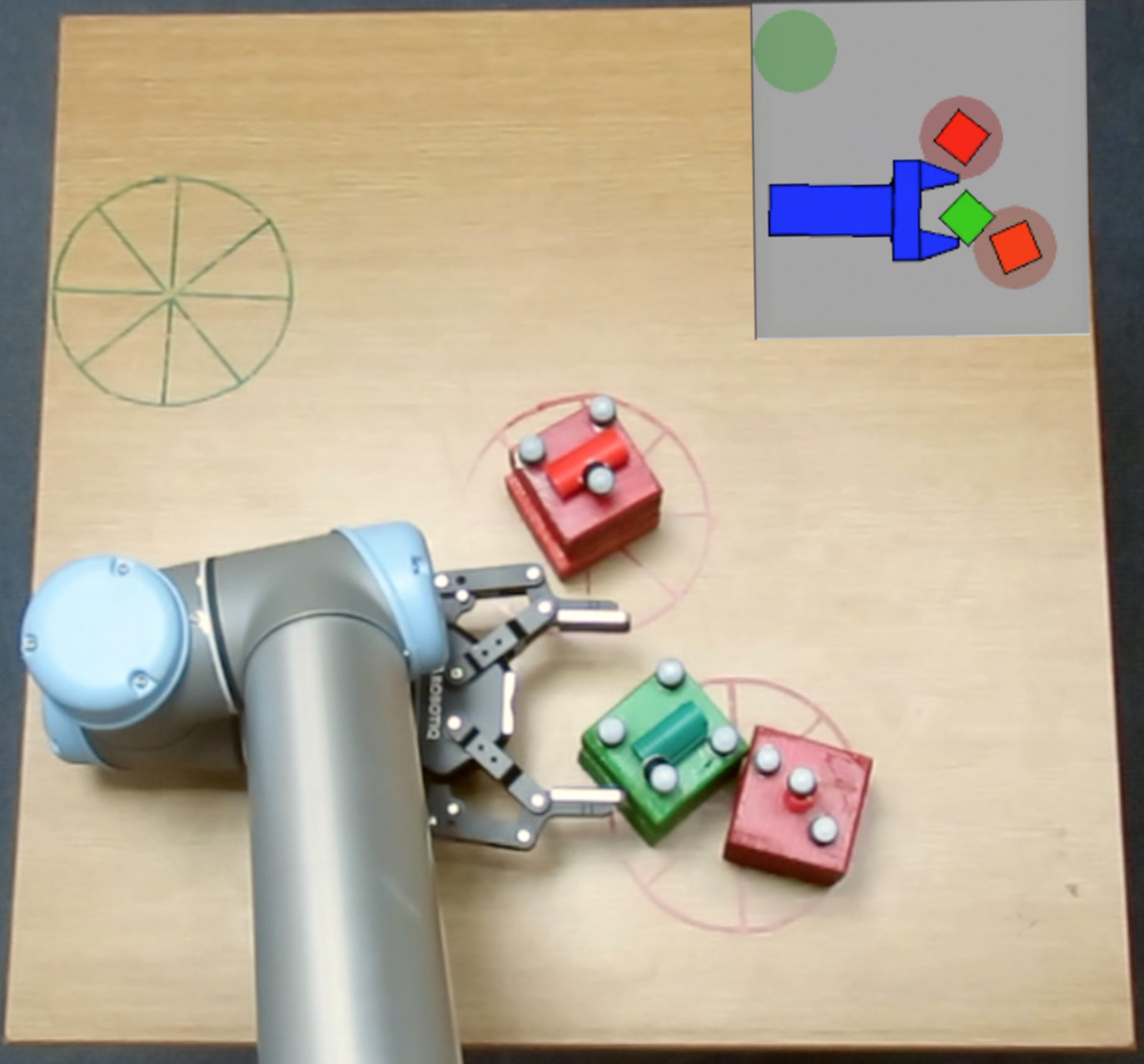}  }\hspace*{-0.92em}
    \subfloat[]{ \includegraphics[width=0.235\columnwidth]{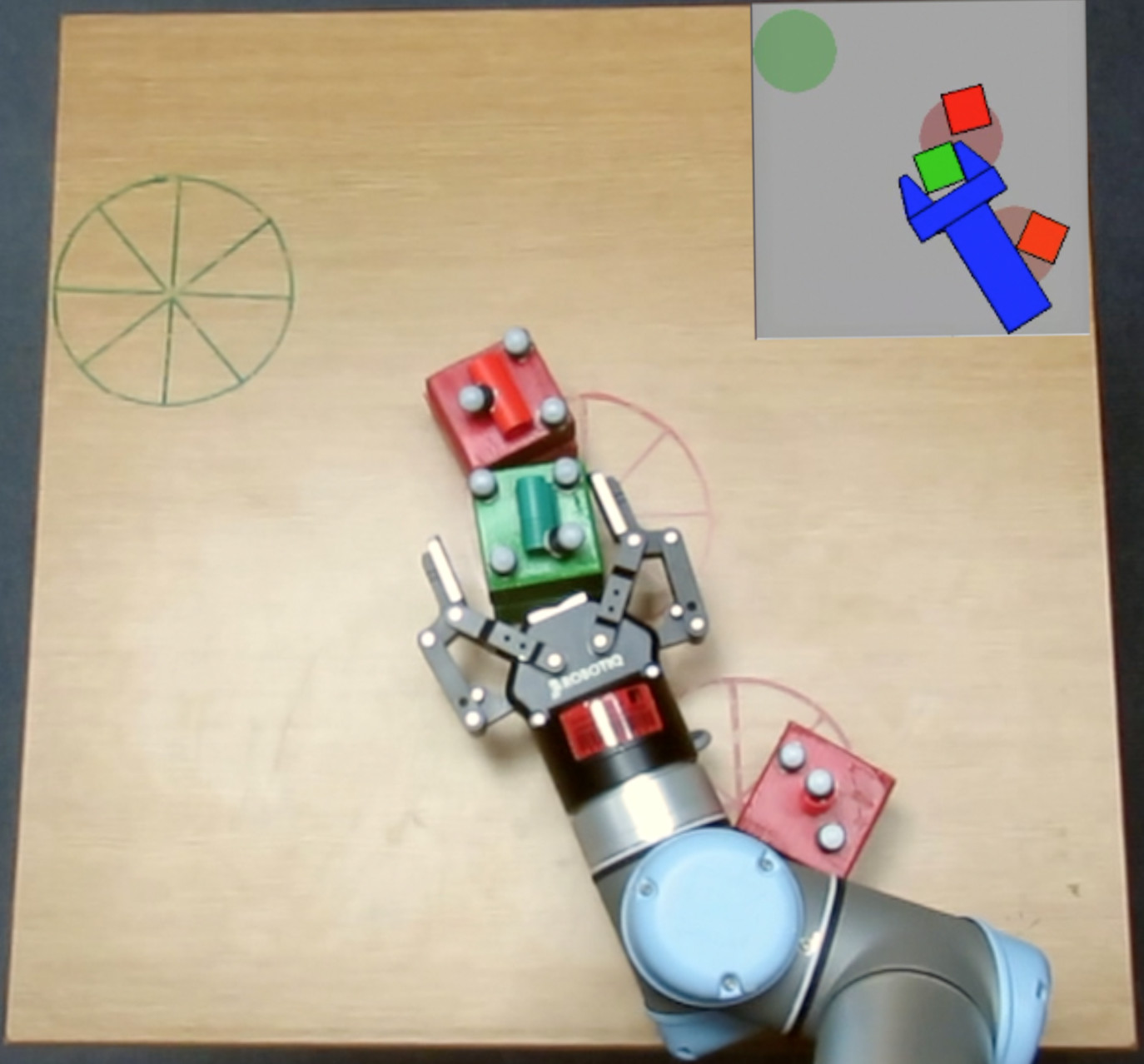}  }\hspace*{-0.92em}
    \subfloat[]{ \includegraphics[width=0.235\columnwidth]{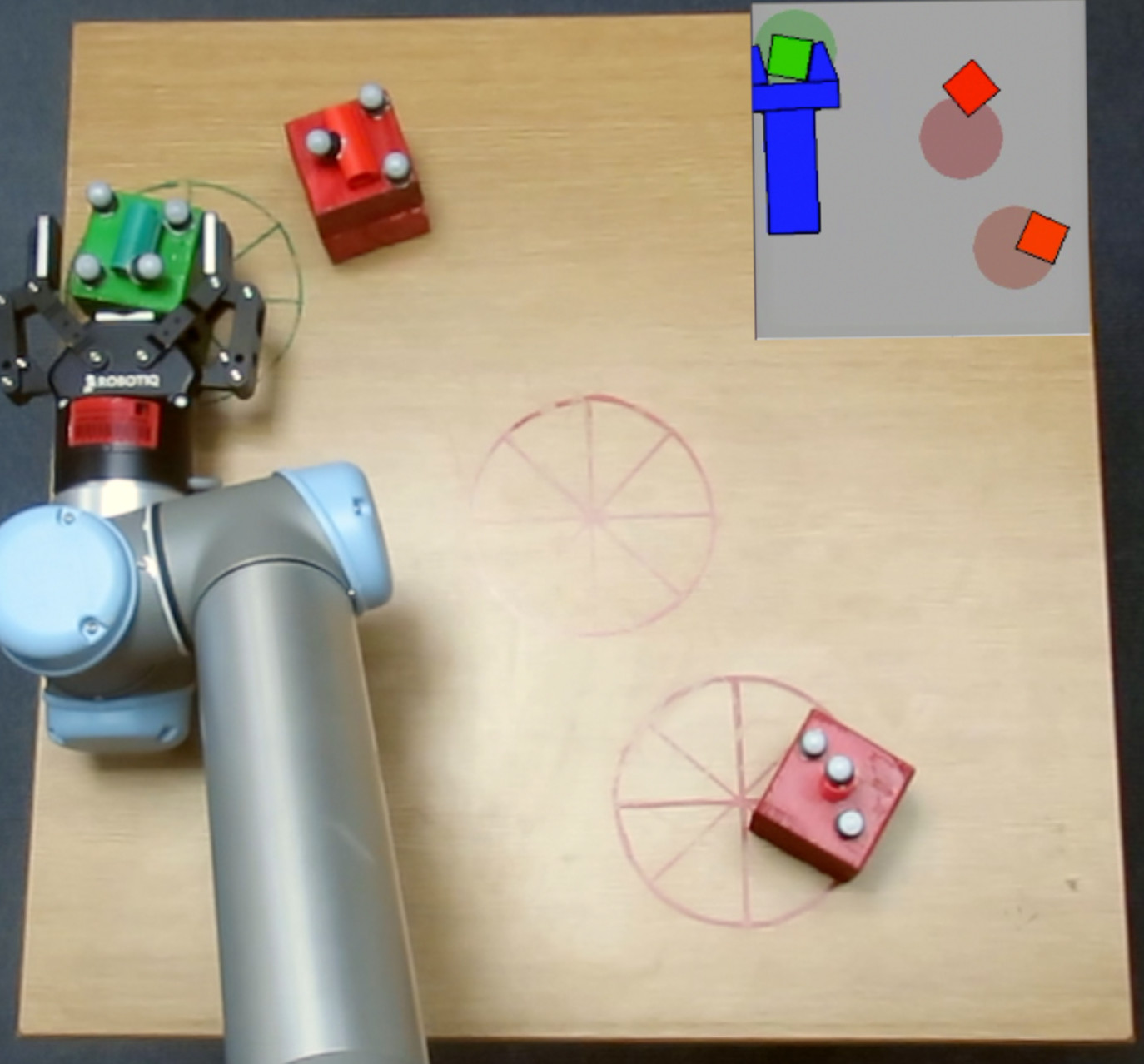}  }\hspace*{-0.92em} \\[-1ex]
    \subfloat[]{\adjustbox{margin=1em,width=0.08\textwidth,set height=0.1cm,angle=90}{Closed-loop}}\hspace*{-0.7em}
    \subfloat[]{\adjustbox{margin=1em,width=0.09\textwidth,set height=0.1cm,angle=90}{RHP execution}}\hspace*{-1.0em}
    \subfloat[]{ \includegraphics[width=0.235\columnwidth]{103po1c.jpeg}  }\hspace*{-0.92em}
    \subfloat[]{ \includegraphics[width=0.235\columnwidth]{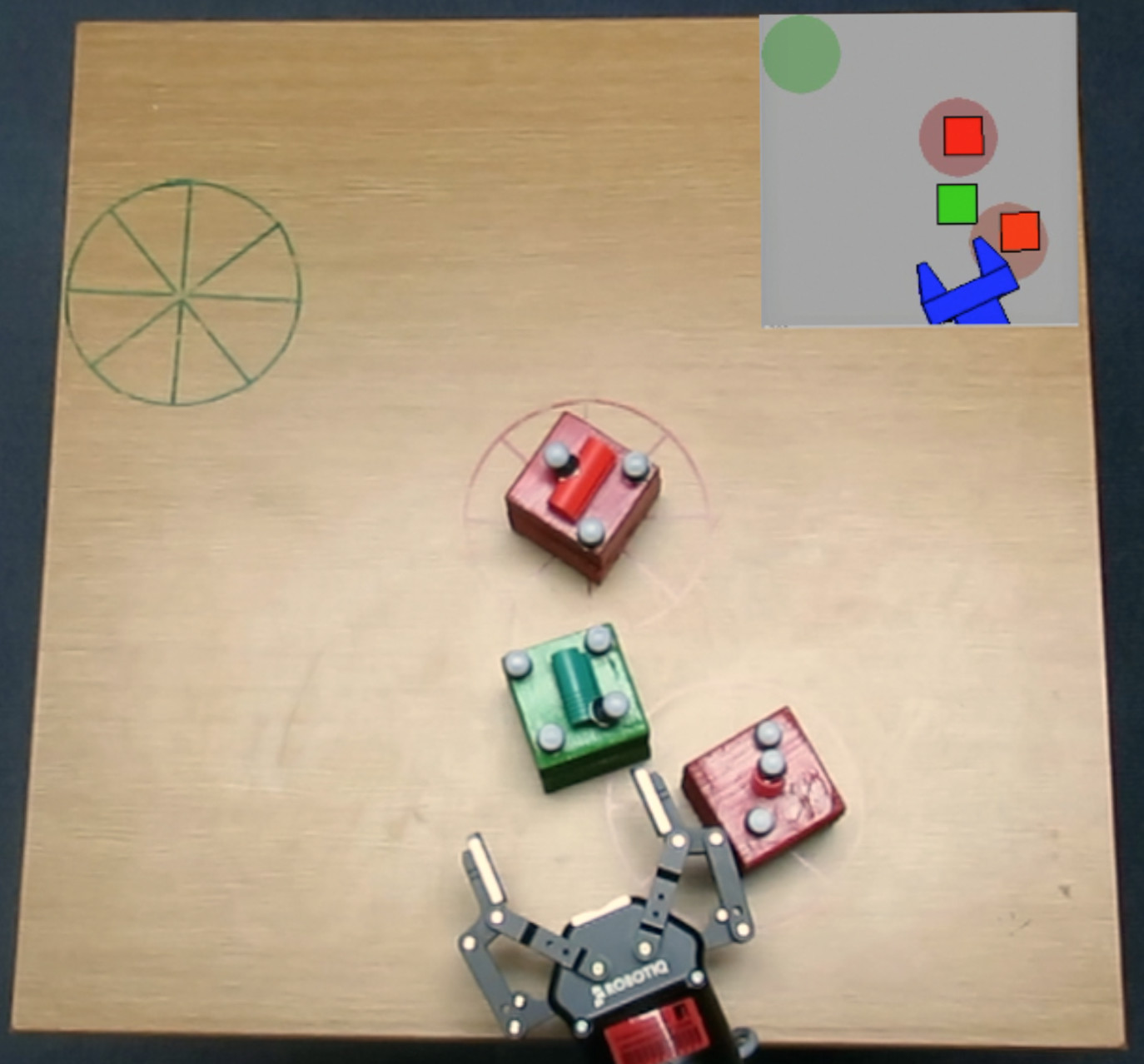}  }\hspace*{-0.92em}
    \subfloat[]{ \includegraphics[width=0.235\columnwidth]{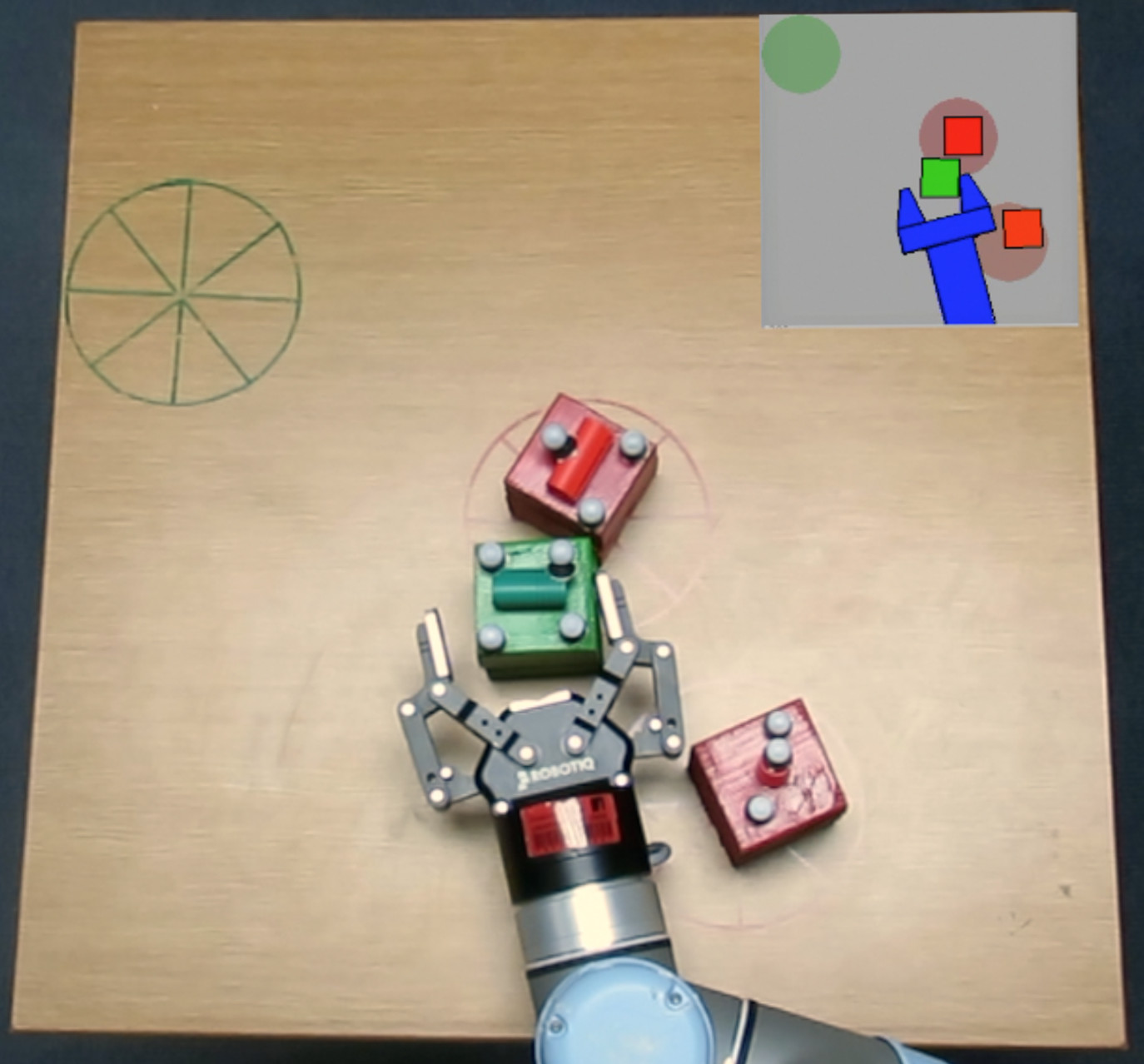}  }\hspace*{-0.92em}
    \subfloat[]{ \includegraphics[width=0.235\columnwidth]{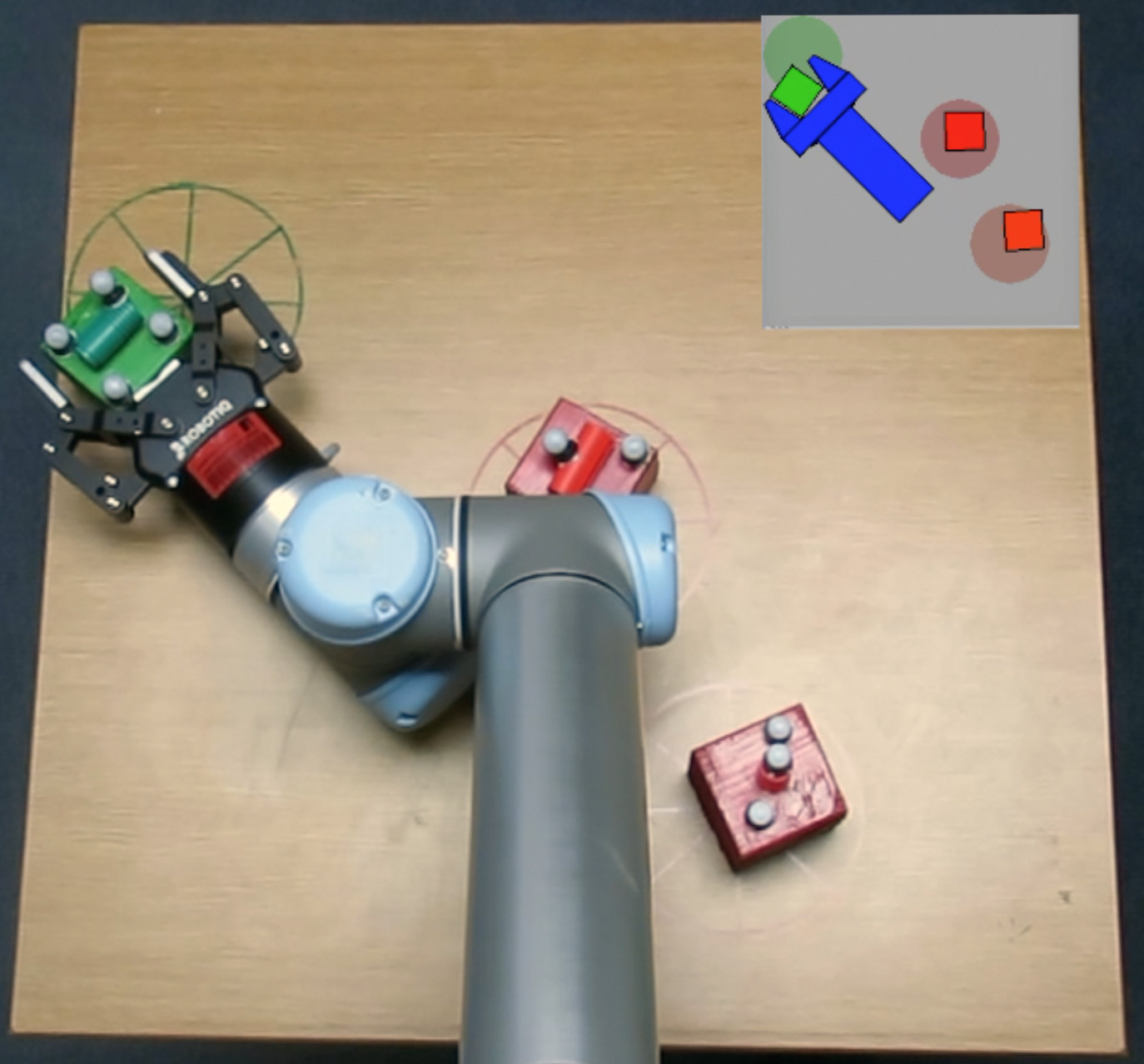}  }\hspace*{-0.92em}
    \caption{Top: robot failing to keep the red box close to its initial position by following a precomputed plan using kino-dynamic planning. Bottom: robot successfully executing the task goal using RHP-66.}
    \label{fig:real_example_2}
\end{figure}

\subsection{Performance evaluation} \label{sec:exp2}

  \begin{table*}[!t]
    \centering
    \caption{The performance results of the different policies and the planner}
    \vspace{5mm}
    \setlength{\tabcolsep}{4pt}
    \begin{tabularx}{\textwidth}{@{\extracolsep{\fill}}lllccclccclccc@{}}
                                           &                         & & \textbf{Planner} & & \multicolumn{3}{c}{\textbf{Kino-dynamic RHP}}  & & \multicolumn{3}{c}{\textbf{Kino-dynamic RHP + RHP-guided RL} } \\ \cmidrule(l){3-12}
                                           &                         & & \textbf{KDP}       & & \textbf{GP}     & \textbf{RHP-33} & \textbf{RHP-66} & & \textbf{GP}    & \textbf{RHP-33} & \textbf{RHP-66} \\
    \midrule
    \multirow{2}{*}{\textbf{No uncert.}}   & suc. rate {[}\%{]}      & & 98.0 $\pm$ 2.0   & & 48.0 $\pm$ 0.0  & 78.0 $\pm$ 1.9     & 88.0 $\pm$ 1.9     & & 51.5 $\pm$ 5.6 & 88.8 $\pm$ 1.3     & 94.4 $\pm$ 1.6     \\
                                           & Avg. exec. time {[}s{]} & & 49.4 $\pm$ 14.9  & & 0.7 $\pm$ 0.0   & 5.8 $\pm$ 0.2      & 17.7 $\pm$ 2.0     & & 0.6 $\pm$ 0.1  & 7.9 $\pm$ 0.2      & 21.2 $\pm$ 1.6     \\ \midrule

    \multirow{2}{*}{\textbf{Low uncert.}}  & suc. rate {[}\%{]}      & & 24.5 $\pm$ 17.3  & & 48.2 $\pm$ 7.7  & 77.2 $\pm$ 4.4     & 86.2 $\pm$ 3.5     & & 48.6 $\pm$ 6.7 & 88.2 $\pm$ 2.8     & 94.0 $\pm$ 2.5     \\
                                           & Avg. exec. time {[}s{]} & & 41.1 $\pm$ 11.7  & & 0.7 $\pm$ 0.1   & 6.4 $\pm$ 0.5      & 18.6 $\pm$ 1.6     & & 0.6 $\pm$ 0.0  & 8.0 $\pm$ 1.3      & 22.7 $\pm$ 5.6     \\ \midrule

    \multirow{2}{*}{\textbf{Med. uncert.}} & suc. rate{[}\%{]}       & & 28.5 $\pm$  25.3 & & 42.2 $\pm$ 12.4 & 73.4 $\pm$ 4.9     & 85.8 $\pm$ 8.7     & & 47.3 $\pm$ 9.1 & 88.0 $\pm$ 2.4     & 91.2 $\pm$ 4.6     \\
                                           & Avg. exec. time {[}s{]} & & 42.5 $\pm$  9.0 & & 0.6 $\pm$ 0.1   & 7.3 $\pm$ 0.1      & 19.7 $\pm$ 3.5     & & 0.6 $\pm$ 0.1  & 8.1 $\pm$ 0.6      & 18.4 $\pm$ 5.2     \\ \midrule

    \multirow{2}{*}{\textbf{High uncert.}} & suc. rate{[}\%{]}       & & 15.7 $\pm$ 15.1  & & 44.7 $\pm$ 29.9 & 71.5 $\pm$ 7.2     & 82.8 $\pm$ 8.1   & & 45.6 $\pm$ 10.4& 87.3 $\pm$ 5.1     & 90.1 $\pm$ 2.8     \\
                                           & Avg. exec. time {[}s{]} & & 37.6 $\pm$ 8.9  & & 0.7 $\pm$ 0.1   & 7.1 $\pm$ 0.8      & 14.5 $\pm$ 2.3  & & 0.7 $\pm$ 0.2  & 8.5 $\pm$ 2.7      & 17.3 $\pm$ 1.6     \\ \bottomrule
    \end{tabularx}
  \label{tab:results}
  \end{table*}

Next, the network that encoded the best action-value function as measured by the  performance in the first round of
experiments is further trained with RHP-guided RL where each RHP query runs $n=6$ roll-outs of $h=6$ horizon depth each.
% \footnote{We found that it was not necessary to use more powerful RL algorithms than DQN, namely DDQN~\cite{van2016deep} and A3C~\cite{mnih2016asynchronous}, because the warm started DNN is already close to convergence.}.
To evaluate the effectiveness of every step in our approach, we compare two
groups of RHP policies:
\begin{itemize}
    \item In the first group, the action-value function (that is, the DNN) is
        learned solely from the plans over the kino-dynamic planner.
        We call this \textbf{kino-dynamic RHP} in Table~\ref{tab:results}.
    \item In the second group, the action-value
        function is further updated with the RHP-guided RL. We call
        this \textbf{kino-dynamic RHP + RHP-guided RL} in Table~\ref{tab:results}.
\end{itemize}

We evaluated each of these groups by using the trained action-value function in
three different ways:
greedy policy (GP), RHP with $n=3, \ h=3$ (RHP-33), and RHP with $n=6, \ h=6$ (RHP-66).
We also include the open-loop execution based on the kino-dynamic planner (KDP)
as a base-line.

\begin{figure}[t] %[thpb]
    \captionsetup[subfigure]{labelformat=empty}
    \setlength{\belowcaptionskip}{-10pt}
%     \centering
	\subfloat[]{\adjustbox{margin=1em,width=0.08\textwidth,set height=0.1cm,angle=90}{Open-loop}}\hspace*{-0.9em}
    \subfloat[]{\adjustbox{margin=1em,width=0.12\textwidth,set height=0.13cm,angle=90}{Kino-dynamic planner}}\hspace*{-1.0em}
    \subfloat[]{ \includegraphics[width=0.235\columnwidth]{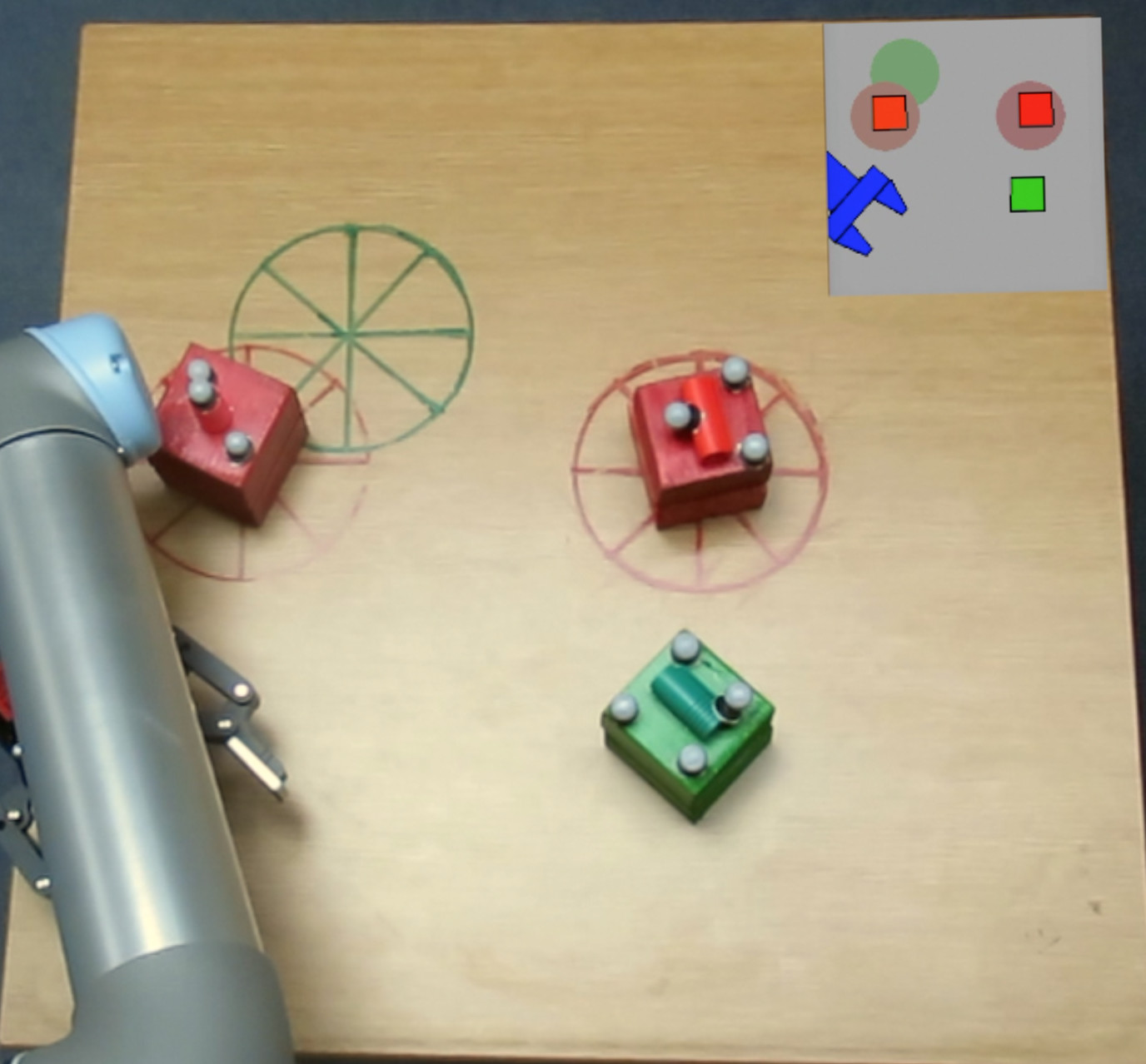}  }\hspace*{-0.92em}
    \subfloat[]{ \includegraphics[width=0.235\columnwidth]{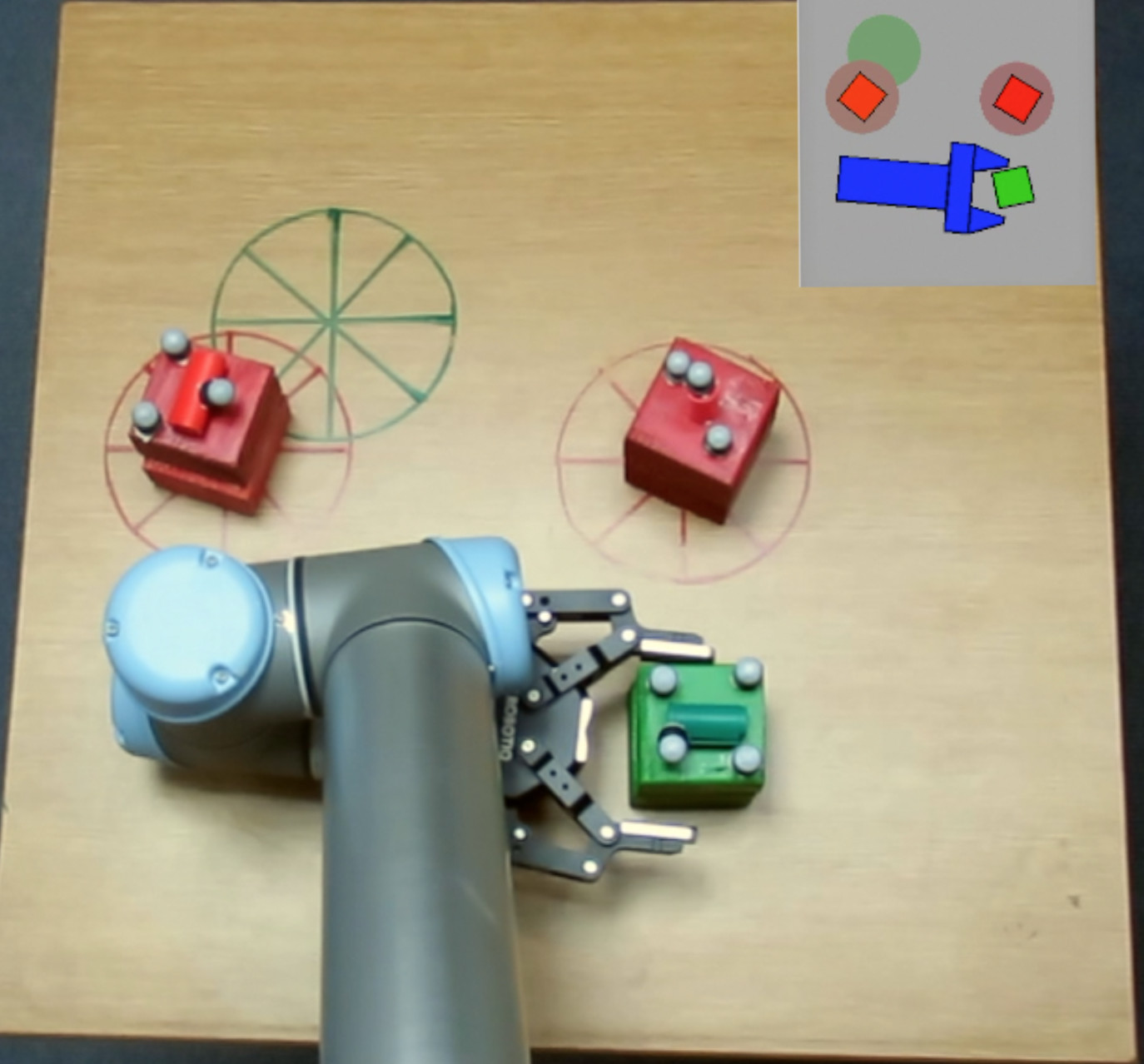}  }\hspace*{-0.92em}
    \subfloat[]{ \includegraphics[width=0.235\columnwidth]{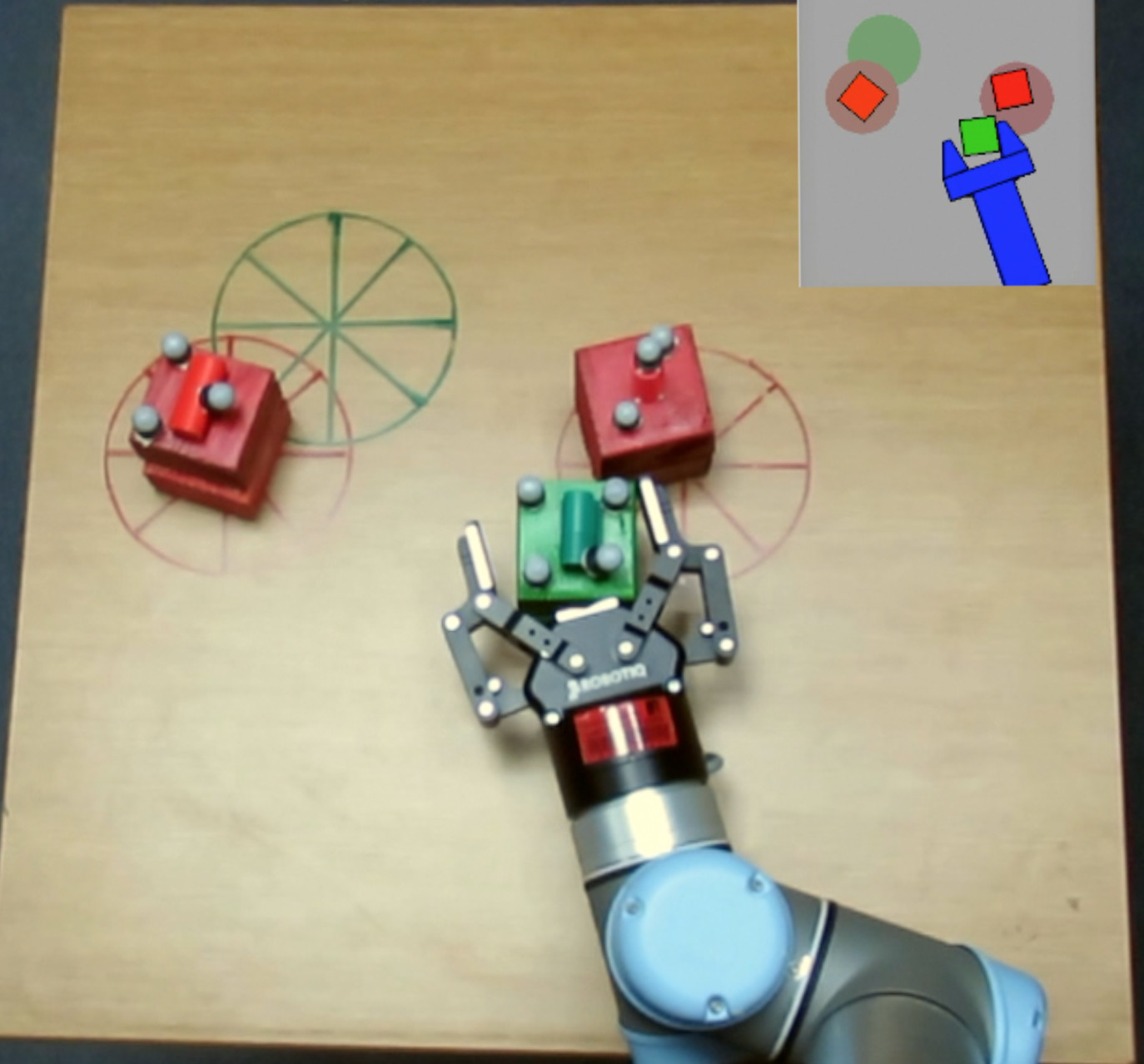}  }\hspace*{-0.92em}
    \subfloat[]{ \includegraphics[width=0.235\columnwidth]{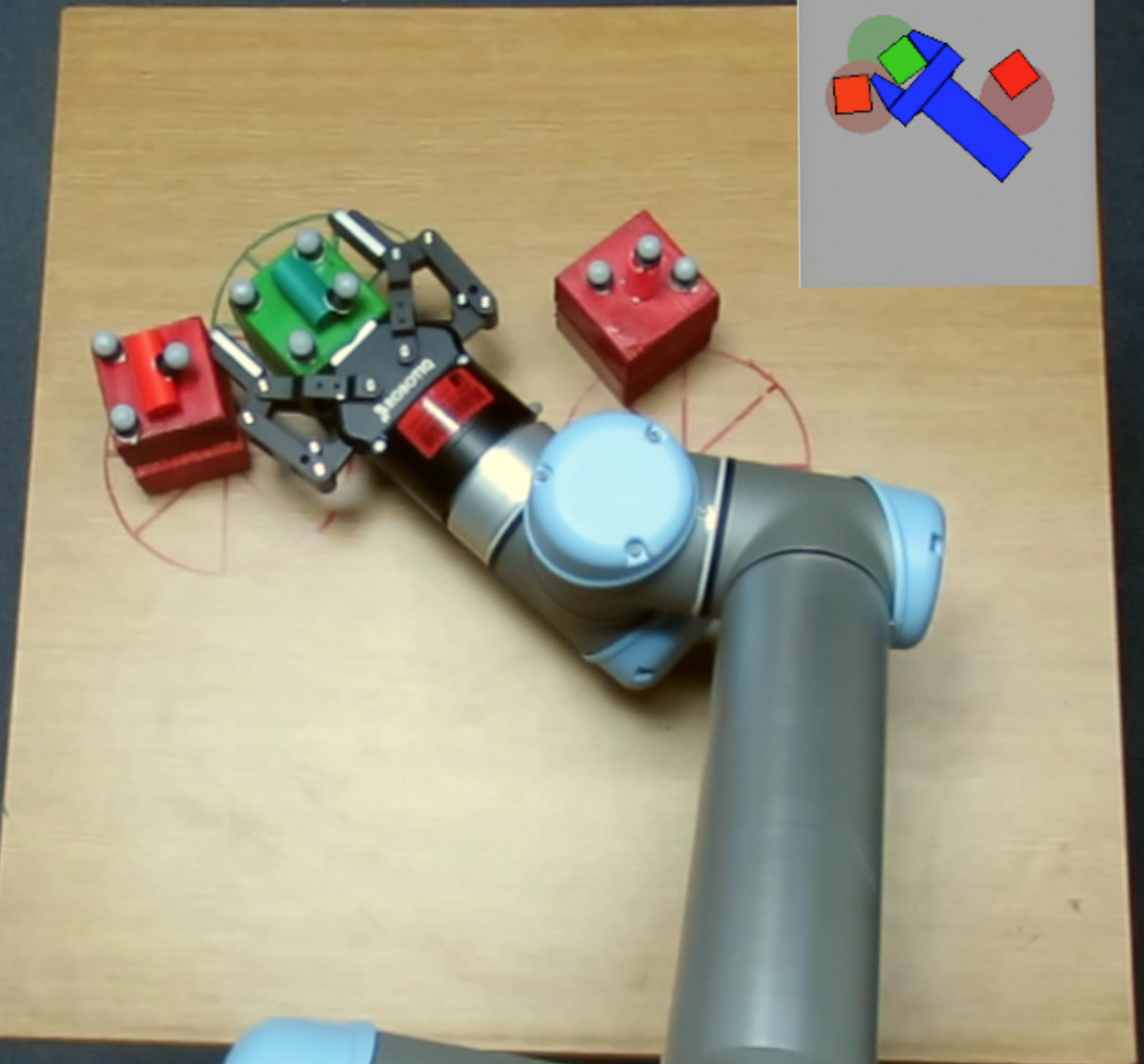}  }\hspace*{-0.92em} \\[-1ex]
    \subfloat[]{\adjustbox{margin=1em,width=0.08\textwidth,set height=0.1cm,angle=90}{Closed-loop}}\hspace*{-0.7em}
    \subfloat[]{\adjustbox{margin=1em,width=0.09\textwidth,set height=0.1cm,angle=90}{RHP execution}}\hspace*{-1.0em}
    \subfloat[]{ \includegraphics[width=0.235\columnwidth]{26po1c.jpeg}  }\hspace*{-0.92em}
    \subfloat[]{ \includegraphics[width=0.235\columnwidth]{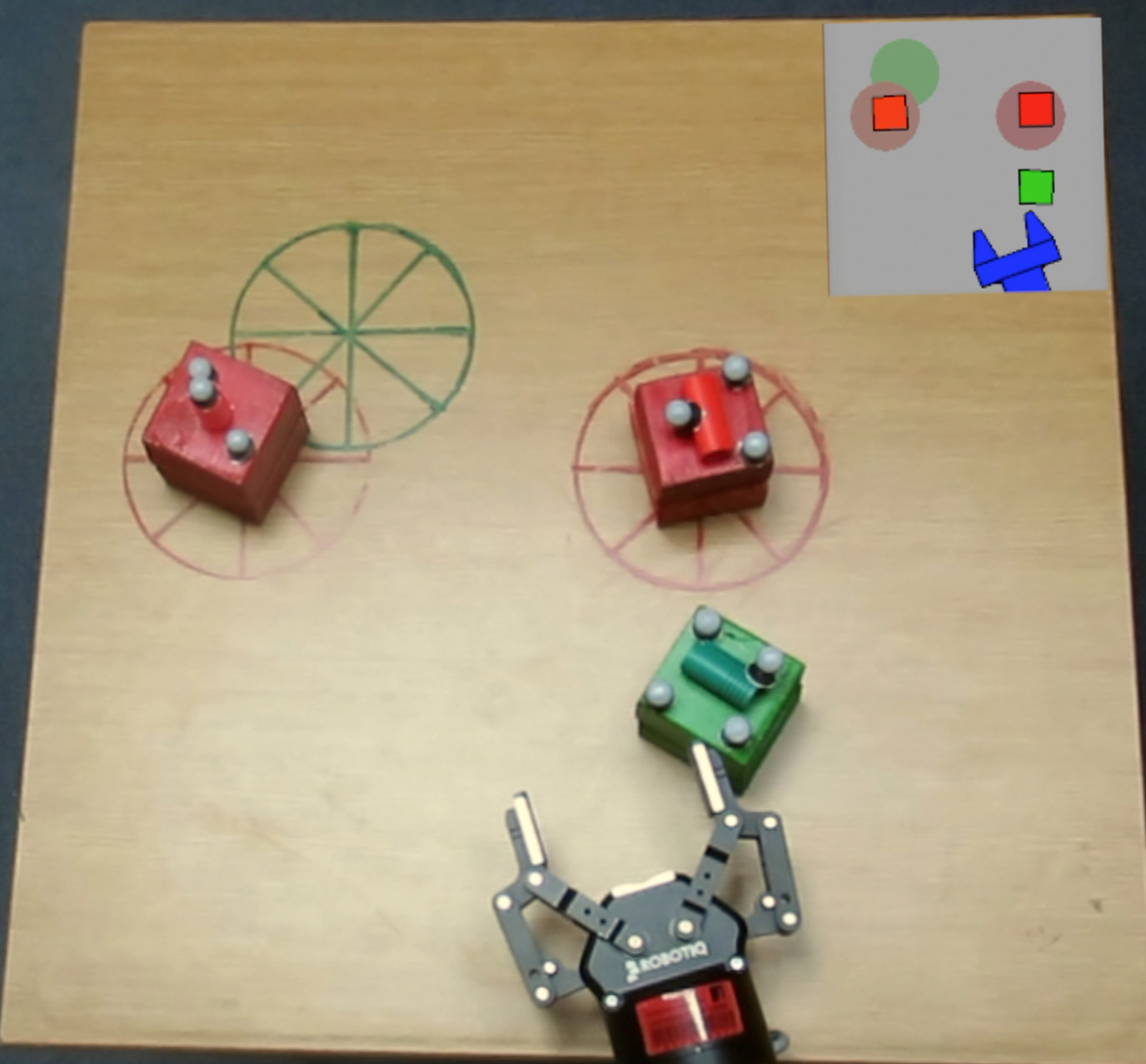}  }\hspace*{-0.92em}
    \subfloat[]{ \includegraphics[width=0.235\columnwidth]{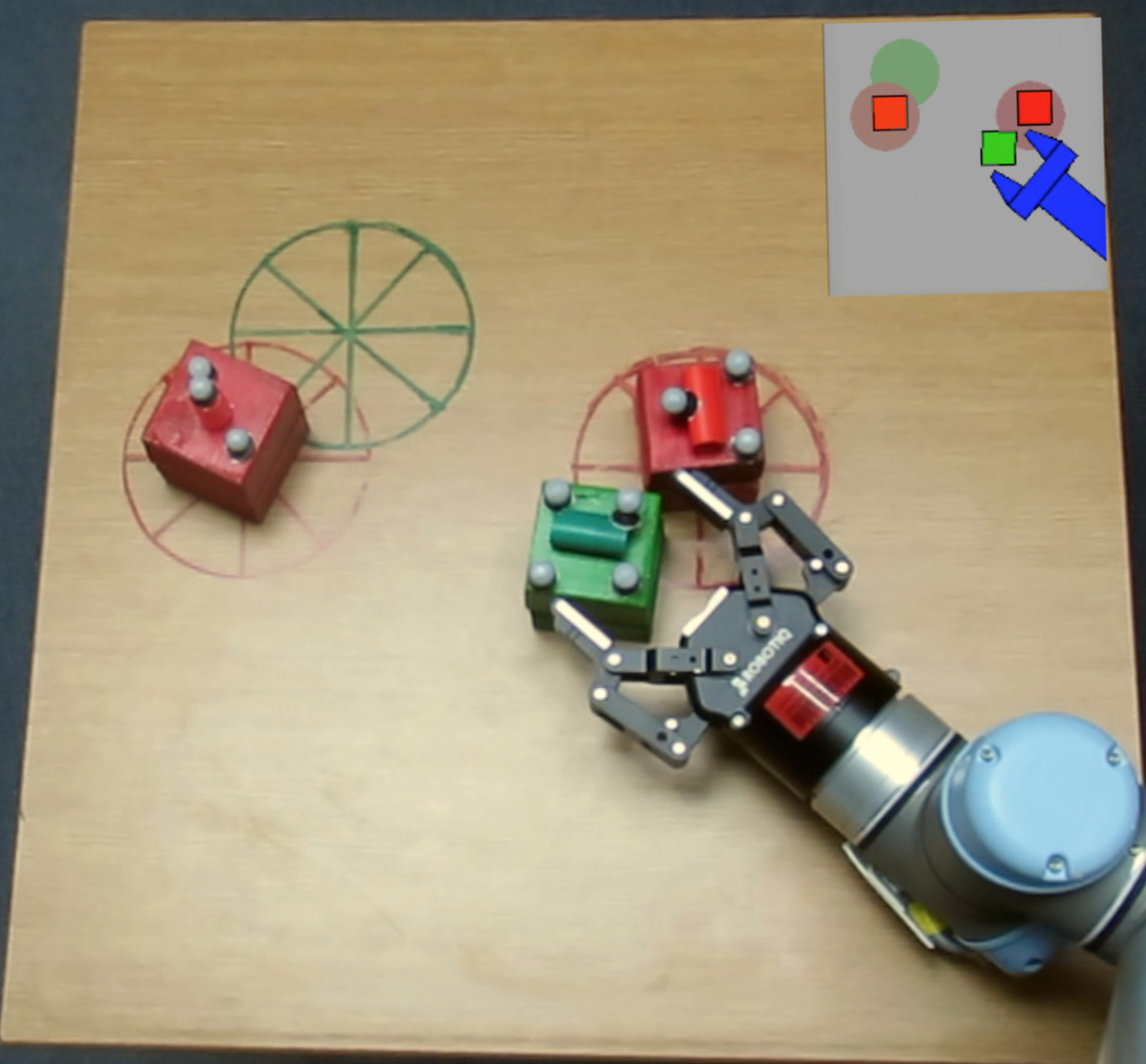}  }\hspace*{-0.92em}
    \subfloat[]{ \includegraphics[width=0.235\columnwidth]{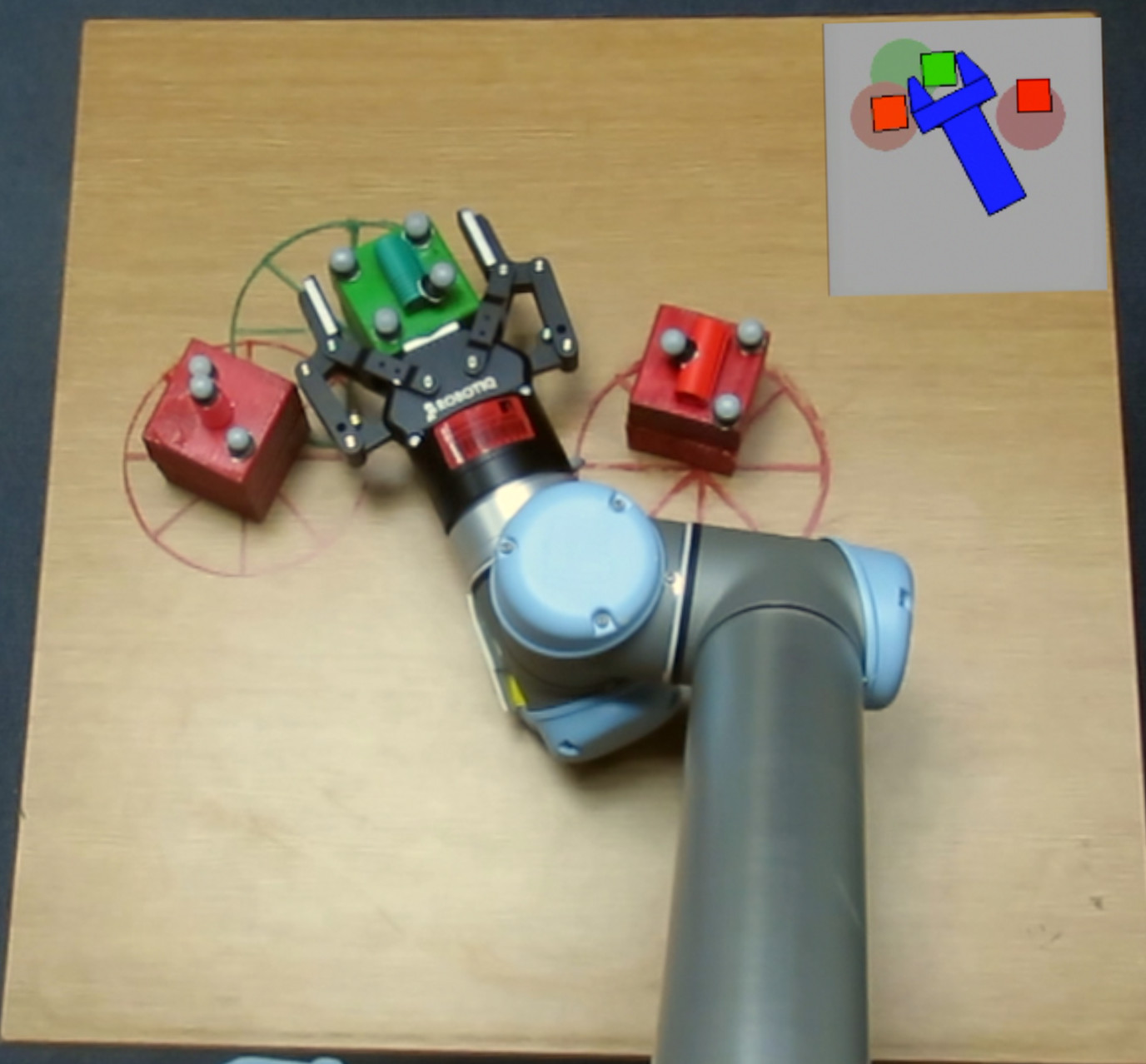}  }\hspace*{-0.92em}
    \caption{Top: robot successfully executing the task goal by following precomputed plan using kino-dynamic planning. Bottom: robot successfully executing the task goal using RHP-66.}
    \label{fig:real_example_3}

\end{figure}

When we evaluated a certain policy,
we injected different levels of uncertainty
in the physics model as a way of gaging how a policy copes with dynamics that are different then the one it was trained on.
The performance under such artificial uncertainty is a way of
estimating the robustness of each policy, and approximating how a policy would perform under real world uncertainty.
The rows in Table~\ref{tab:results} correspond to these
uncertainty levels.

To inject uncertainty into the execution, we considered physics parameters:
shape, friction, and density of the boxes. During evaluation, the uncertainty is sampled from a Gaussian
distribution centered around the value of the parameters used in the
training (and planning in the kino-dynamic planner case)\footnote{\emph{Mean} values of the boxes' physics parameters: shape:0.06x0.06$m$, density: $1$ $kg/m^2$, friction coefficient: $0.3$. \\
Standard deviation on the boxes' physics parameters with corresponding uncertainty: Low = $0.1 \times mean$, medium = $0.2 \times mean$, high = $0.3 \times mean$.}.

In each cell, Table~\ref{tab:results} shows
the success rate and the average computation time per successful
execution. The latter includes planning time, whether kino-dynamic planning
or RHP, and the time required to compute the physical interaction
using Box2D. Also, a computation time limit of 2 minutes is imposed on all trials.
The results presented are averaged over
10 trials on the 500 random task instances. We note that the experiments are conducted on
an Intel Xeon E5-2665 computer equipped with NVIDIA Quadro K4000 GPU card.

The kino-dynamic planner case with no uncertainty shows a high success rate. The
few cases where it failed are due to the imposed time limit. Nevertheless, the
decreasing performance with uncertainty and the relatively high computation
time confirms the limitation of using open-loop planning in execution. The left image in Fig.
\ref{fig:sim_exp} shows how a plan can fail during execution when the
environment is slightly different than expected. The green box was expected to slide
inside the robot hand, however because of mismatched dimension, that is, the box has a
rectangular shape instead of the square shape used for planning, the box slides
outside of the arm trajectory. In general, this indicates that this type of
planning is favorable when a high-fidelity model and high-processing power are available.

We also notice that when RHP is engaged there is a notable
increase in performance. The longer the horizon and number of roll-outs the
higher is the success rate and the more robust it is against uncertainty.
The performance increase comes at a cost of an increased computation time.
However, it
is still within reasonable limits for near real-time manipulation. In
contrast to using an open-loop control scheme, the right image in Fig. \ref{fig:sim_exp}
illustrates how the robot can adapt to unexpected behaviors.

Looking at the overall performance between the two groups of policies, we see
that further optimizing the action-value function with RHP-guided RL contributed to a
higher success rate and robustness to uncertainty.
Particularly, RHP-66 outperformed all of the others \wrt the success rate.
We used this policy successfully to command a robot in the real world.

% \commentw{We note that we also compared our results to a state-of-the-art RL algorithm, namely the DDQN algorithm \cite{van2016deep}. The algorithm did not yield to the convergence of the action-value function. Hence, it was not possible to induce a policy from it for controlling the robot. This can be attributed to the sparse reward nature of our pushing task which makes it highly unlikely for the robot to reach the goal for the first time without any reward shaping.}

\subsection{Real robot execution}

We performed experiments on a UR5 robot\footnote{https://www.universal-robots.com/products/ur5-robot/}.
We created the three task instances shown in
Figures~\ref{fig:real_example_1},~\ref{fig:real_example_2},~and~\ref{fig:real_example_3}.
In each task, we tested the trained RHP-66 policy (bottom row in figures) and
compared it to the open-loop execution of the kino-dynamic planner (top row).
During the execution of RHP, closed-loop feedback on object poses was
supplied using an OptiTrack system for RHP to run the roll-outs on the model.
As expected, the reactive capability of
RHP made its reaction robust to the dynamics of the real world, and succeed in these three tasks.
In two out of three tasks, the open-loop execution failed.
A video of these experiments is available on \textcolor{myblue}{https://youtu.be/xwa0fTTuQ1g}.

\section{Conclusions} \label{sec:sum}
This paper described a receding horizon planning (RHP) approach for
closed-loop planning to solve
physics-based manipulation in clutter problems in
near real-time.
We demonstrated how a suitable
action-value function for RHP can be learned
from a sampling-based planner,
and how further improving the action-value function with RL
contributes to the system being both faster, and more robust,
than the open-loop planner it builds on.
Our approach does not require engineering domain-dependent
heuristics or manual reward shaping.

These findings motivate us to further develop our research on real-time dynamic
manipulation. We are currently extending this work to admit visual input for the state representation. This will allow the robot to seamlessly adapt to a changing number of objects in the scene.
We are also committed to augmenting the robot manipulation skills with other manipulation primitives such as grasping, leveraging, and rolling.

\bibliographystyle{IEEEtran} %plain %abbrv %IEEEtran
\bibliography{mybib}

\addtolength{\textheight}{-12cm}   % This command serves to balance the column lengths

\end{document}